%% file: iclr2027_conference.tex
\documentclass{article}
\usepackage{iclr2027_conference,times}
\iclrfinalcopy

\input{math_commands.tex}
\input{0-preamble}

\usepackage{cleveref}

\input{appendix/macros}

\title{\CAPEX: Efficiently Distilling Foundation Model Behavior into Deployable Robot Policies through Experience-Adaptive Reasoning}

\author{Shivam Aarya, Zhang Xi-Jia, Chengyue Huang, Junhyun Kim, Huishu Xue, \\
\textbf{Hrishit Leen, Roman Yakunin, Animesh Garg, Zsolt Kira} \\
School of Interactive Computing \\
Georgia Institute of Technology \\
Atlanta, GA 30332, USA
}

\begin{document}

\maketitle
\lhead{Preprint. Under review.}

\input{sections/0_abstract}

\input{sections/1_intro}

\input{sections/2_related_work}
\input{sections/3_method}
\input{sections/4_exp}
\input{sections/5_conclusion}

\subsubsection*{Acknowledgments}
This work was supported by API research credits from OpenAI.

\bibliography{iclr2027_conference}
\bibliographystyle{iclr2027_conference}

\newpage
\appendix
\input{appendix/appendix}

\end{document}

%% file: math_commands.tex
\usepackage{amsmath,amsfonts,bm}

\def\eqref#1{equation~\ref{#1}}

\def\1{\bm{1}}

\DeclareMathAlphabet{\mathsfit}{\encodingdefault}{\sfdefault}{m}{sl}
\SetMathAlphabet{\mathsfit}{bold}{\encodingdefault}{\sfdefault}{bx}{n}

\newcommand{\CAPEX}{\textsc{CAPEX}\xspace}

%% file: 0-preamble.tex
\usepackage[T1]{fontenc}
\usepackage[utf8]{inputenc}
\usepackage{microtype}
\usepackage{fix-cm}
\usepackage{comment}

\PassOptionsToPackage{table}{xcolor}
\usepackage{xcolor}
\usepackage{colortbl}
\usepackage{booktabs}
\usepackage{array}
\usepackage{tabularx}
\usepackage{multirow}
\usepackage{multicol}
\usepackage{tablefootnote}
\usepackage{adjustbox}
\usepackage{soul}

\definecolor{codegreen}{rgb}{0,0.6,0}
\definecolor{codegray}{rgb}{0.4,0.4,0.4}
\definecolor{codepurple}{rgb}{0.5,0,0.9}
\definecolor{backcolour}{rgb}{0.95,0.95,0.95}
\definecolor{lightgray}{rgb}{0.9,0.9,0.9}
\definecolor{lightpink}{rgb}{0.98,0.85,0.86}
\definecolor{lightblue}{rgb}{0.68,0.84,0.9}

\definecolor{accentBlue}{HTML}{DDE8F2}
\definecolor{accentYellow}{HTML}{F3E6C9}
\definecolor{accentPurple}{HTML}{E9DFF2}
\definecolor{accentGreen}{HTML}{D9EDE1}

\renewcommand{\arraystretch}{1.1}
\newcolumntype{Y}{>{\raggedright\arraybackslash}X}

\usepackage{amsmath,amssymb,mathtools}
\usepackage{bm}
\usepackage{bbm}
\usepackage{nicefrac}
\usepackage{textcomp}
\usepackage{pifont}
\usepackage{xspace}
\usepackage[
  separate-uncertainty = true,
  multi-part-units = repeat
]{siunitx}

\usepackage{graphicx}
\usepackage{float}
\usepackage{subcaption}
\usepackage[font=small,labelfont=bf]{caption}
\usepackage{lscape}
\usepackage{tikz}
\usetikzlibrary{arrows.meta}
\usepackage{makecell}
\usepackage{overpic}

\usepackage{algorithm}
\usepackage[noend]{algpseudocode}

\usepackage{paralist}
\usepackage{enumitem}
\setlist[itemize]{noitemsep,leftmargin=*,topsep=0in}
\setlist[enumerate]{noitemsep,leftmargin=*,topsep=0in}

\usepackage{url}
\usepackage{hyperref}
\hypersetup{
  colorlinks=true,
  citecolor=blue!45!black,
  linkcolor=blue!45!black,
  urlcolor=blue!45!black,
  bookmarks=false
}

\usepackage{listings}
\lstdefinestyle{mystyle}{
    backgroundcolor=\color{backcolour},   
    commentstyle=\color{codegreen},
    keywordstyle=\color{magenta},
    numberstyle=\tiny\color{codegray},
    stringstyle=\color{codepurple},
    basicstyle=\fontsize{8}{8}\selectfont\ttfamily,
    breakatwhitespace=false,         
    breaklines=true,
    breakindent=0pt,
    captionpos=b,                    
    keepspaces=true,                 
    numbers=none,                    
    numbersep=5pt,                  
    showspaces=false,                
    showstringspaces=false,
    showtabs=false,                  
    tabsize=2
}
\usepackage{titlesec}
\titlespacing{\section}{0pt}{0.3\baselineskip}{0.25\baselineskip}
\titlespacing{\subsection}{0pt}{0.2\baselineskip}{0.15\baselineskip}
\titlespacing{\subsubsection}{0pt}{0.05\baselineskip}{0.03\baselineskip}
\renewcommand{\paragraph}[1]{\vspace{0.2em}\noindent\textit{#1} --}

\usepackage{wrapfig}
\usepackage{placeins}

%% file: appendix/macros.tex
\providecommand{\ours}{CAPEX}
\providecommand{\astra}{GPT-6 Astra}
\providecommand{\qwen}{Qwen3.8-27B}

%% file: sections/0_abstract.tex
\begin{abstract}

Robot learning has largely relied on human-teleoperated demonstrations to acquire effective learnable behaviors. However, human-operated data collection processes can be unintuitive, difficult to scale, and inherently asynchronous. We explore an alternative: distilling physical behavior from general-purpose multimodal foundation models into deployable robot policies by using the foundation model itself as an autonomous demonstrator. While sufficiently capable models can generate successful zero-shot manipulation trajectories, repeatedly invoking them during physical execution is slow and expensive, limiting their utility as scalable data generators. 
As a solution, we introduce \CAPEX, an experience-conditioned demonstration collection framework that uses execution experience from previous attempts to adapt how frequently the foundation model must observe, reason, and replan. We evaluate across RoboCasa tasks and on physical Franka and bimanual YAM-arm platforms, measuring task success, model calls, token usage, collection time, and cost. We further train Diffusion Policy and ACT on matched sets of human-teleoperated and foundation-model-generated demonstrations to evaluate the downstream learning value of autonomously collected data. We find that \CAPEX increases the number of successful demonstrations by \(4.3\times\) while reducing the cost per successful demonstration by \(80\%\). %
Policies trained on \CAPEX-generated data approach the performance of those trained on matched human demonstrations; with longer training, this gap largely closes for policies trained from scratch. These results suggest that foundation models can serve as scalable sources of reusable robot experience.

Project page: \url{https://capex-paper.github.io/}

\end{abstract}

%% file: sections/1_intro.tex
\section{Introduction}

Learning effective robot policies has historically relied on human-teleoperated demonstrations \citep{zhao2023act,chi2023diffusion,zitkovich2023rt2,ghosh2024octo,kim2024openvla} that are bottlenecked by human effort: teleoperation can be difficult to operate, demonstrations must be collected through dedicated interaction, and scaling data requires scaling the corresponding human labor \citep{qin2023anyteleop,wu2023gello,zhao2023act,chi2024umi}. This bottleneck has motivated methods that amplify limited demonstrations or increasingly automate robot data collection \citep{mandlekar2023mimicgen,irpan2024autort,liang2026tether}. Now, increasingly capable general-purpose foundation models offer a new source of supervision \citep{openai2026gpt6astra, anthropic2026fable51}. Recent frontier multimodal models such as GPT-6 Astra suggest a qualitative shift in the role of general-purpose intelligence in robotics. These models remove the need for any human demonstrations and directly generate executable behavior for real-world embodied systems, producing successful rollouts across settings including bimanual manipulation \citep{zhang2026unexpected}, mobile manipulation \citep{pantograph2026benchmarking}, and autonomous driving \citep{drivingbench2026astra}. This emerging capability raises a different possibility for robot learning: instead of relying on humans to directly generate every robot trajectory, can we use general intelligence to autonomously generate physical experience and then distill that experience into deployable robot policies?

\begin{figure}[t]
  \centering
  \includegraphics[width=\textwidth]{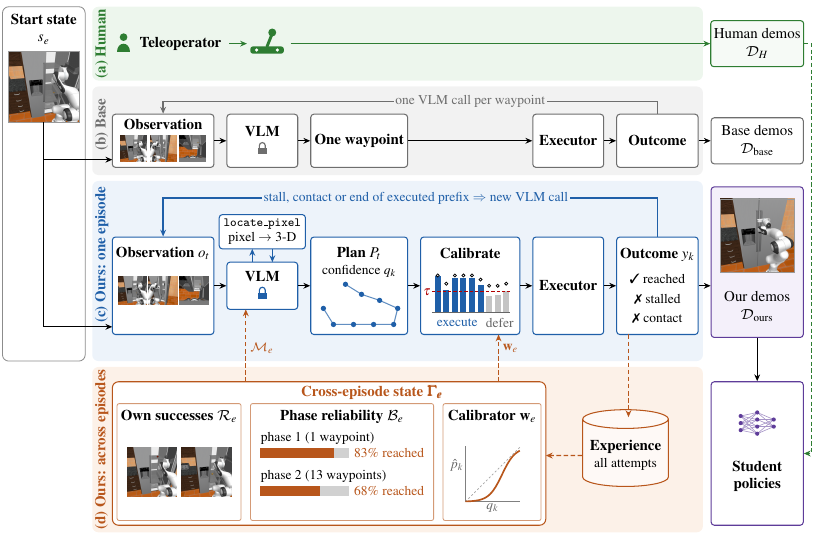}
  \caption{\textbf{\CAPEX overview.}
  All arms start from the initial state $s_e$ of a human demonstration (CloseFridge shown).
  (a)~Teleoperation yields $\mathcal{D}_H$.
  (b)~The base arm queries a VLM once per waypoint.
  (c)~\CAPEX's VLM proposes a multi-waypoint plan with stated confidences $q_k$ (circles); the robot
  executes it until a later waypoint's calibrated probability $\hat p_k$ (bars) falls below $\tau$ or
  execution stops, then queries the VLM again.
  (d)~Before each episode, all past attempts are summarized into $\Gamma_e$: calibrator weights
  $\mathbf{w}_e$, which set how much of a plan runs, and a VLM context $\mathcal{M}_e$, which shapes the plan
  with the robot's own successes $\mathcal{R}_e$ and how reliably each task phase has executed $\mathcal{B}_e$.
  Student policies are trained on $\mathcal{D}_{\text{ours}}$ and $\mathcal{D}_H$. Dashed arrows cross episodes.}
  \vspace{-5pt}
  \label{fig:overview}
\end{figure}

We study this alternative paradigm, in which a general-purpose multimodal foundation model acts as an autonomous robot demonstrator, and successful demonstrations become training data for downstream policies. The key challenge is that repeatedly invoking a frontier model throughout physical execution is itself slow and expensive, limiting its usefulness as a accessible source of robot training data.
As a solution, we introduce \textbf{C}alibrated \textbf{A}daptive \textbf{P}lanning through \textbf{EX}perience Collection (CAPEX), an experience-adaptive framework for efficiently collecting demonstrations with a multimodal foundation model (see Figure \ref{fig:overview}). \CAPEX reduces the cost of obtaining each successful demonstration by learning when expensive reasoning is actually needed.
Rather than querying the model after every small motion, \CAPEX allows it to propose temporally extended waypoint plans and learns from physical execution how much of those plans can be executed before reasoning again. 
Using physically self-supervised confidence calibration and bounded context from the robot's own previous attempts, \CAPEX assigns a confidence score to each waypoint, allowing it to commit to confident plan prefixes while returning to the model when later motions are uncertain.
At each decision, the foundation model proposes a variable-length waypoint plan together with a reach probability for each waypoint. Execution outcomes from previous attempts recalibrate these probabilities, allowing \CAPEX to execute within a confident horizon while returning to the model when later motions are uncertain.

Across RoboCasa and physical Franka and bimanual YAM platforms, we evaluate whether foundation models can generate successful zero-shot manipulation trajectories, and further, whether they can do so efficiently enough to serve as practical data generators. \CAPEX increases successful demonstration collection from 12.5\% to 53.6\%, while reducing foundation-model requests per successful demonstration by 84.8\% (244 to 37), collection time by 83.4\% (39.7 to 6.6 minutes), and API cost by 80.0\% (\$9.15 to \$1.83). Finally, we train downstream imitation policies on
matched quantities of human-teleoperated and foundation-model generated data to measure the learning value of the resulting demonstrations. With longer training, policies learned from scratch largely close the gap to those trained on human demonstrations, and policies trained on autonomously collected data remain effective on physical manipulation tasks. Together, these experiments evaluate foundation models not only as online robot controllers, but as scalable generators of reusable robot training experience.

%% file: sections/2_related_work.tex
\section{Related Work}
\label{sec:related_work}

\subsection{Autonomous data generation for robot policies.}

Prior work reduces human collection effort by amplifying sparse demonstrations through trajectory transformation \citep{mandlekar2023mimicgen,jiang2025dexmimicgen,ameperosa2025rocoda,liang2026tether}, generating tasks and supervision in simulation \citep{wang2024gensim,wang2024robogen,nasiriany2026robocasa365}, or transferring supervision from human embodiment data \citep{kareer2025egomimic,dan2025xsim}. 
Foundation-model-based systems further automate interaction, but often use the model to provide task decomposition, spatial constraints, or programs while separate planners or controllers generate executable actions \citep{liang2023code,singh2023progprompt,huang2023voxposer,ha2023scaling}.
Manipulate-Anything similarly targets autonomous demonstration generation, using VLMs for task decomposition, action generation, and execution verification, while leveraging separate grasp prediction and motion-planning components \citep{duan2024manipulate}. \CAPEX instead studies the emerging regime in which a general-purpose multimodal model is repeatedly queried throughout execution to directly propose geometric robot actions from observations. This more directly leverages the model's visuospatial reasoning for closed-loop behavior, but makes frontier-model inference a dominant collection cost. \CAPEX addresses this bottleneck by using execution experience to adapt how much of each proposed action sequence can be executed before the model must reason again.

\subsection{Foundation-model for robot control and execution.}
Foundation models enter robot control both as learned vision-language-action policies \citep{zitkovich2023rt2,ghosh2024octo,kim2024openvla} and as semantic reasoners connected to robot skills, feedback, code, or motion planners \citep{ahn2022saycan,huang2022inner,liang2023code,huang2023voxposer}; hierarchical systems similarly separate deliberative vision-language reasoning from fast action generation \citep{nvidia2025groot,shi2025hirobot}. Concurrent work directly deploys general-purpose frontier VLM agents as robot policies through structured tools and in-context interaction \citep{cheng2026gptpolicy,jia2026agentaspolicy}; Agent-as-Policy further measures wall-clock time, token usage, and API cost per physical execution, highlighting repeated frontier-model inference as a practical cost of agentic robot control. 
Prior experience has been incorporated through self-generated training data, action correction, retrieved memory, and reusable skill libraries \citep{bousmalis2024robocat,skreta2023errors,xiong2024aic,byrnes2025climb,li2026mapvla,lu2026aspire}.

\looseness=-1 Temporally extended action prediction reduces effective decision horizon but trades temporal consistency against closed-loop reactivity \citep{zhao2023act,chi2023diffusion,liu2025bid}, motivating adaptive execution based on phase structure, prediction uncertainty, learned continue-or-replan decisions, or execution progress \citep{nie2026pace,feng2026dvac,xu2026bcp,ye2026tempowam}. Related work also allocates computation or explicit reasoning according to decision difficulty \citep{yue2024deer,lin2026onetwovla,liu2026rarrl,cai2026tau0}, reuses or overlaps computation across policy evaluations \citep{xu2025vlacache,black2025rtc}, or periodically re-queries a frozen frontier model while amortizing calls through caching and reusable control code \citep{naouali2026vlcp}. \CAPEX combines these concerns in autonomous data collection: physical execution outcomes self-supervise confidence calibration and selective plan commitment, while bounded phase summaries and successful prior episodes condition future planning without updating the foundation model.

%% file: sections/3_method.tex
\section{\CAPEX: VLM-Guided Scalable Demonstration Collection}
\label{sec:method}

We study how a frozen multimodal foundation model can be used as an autonomous source of robot demonstrations. For task instruction $\ell$ and start state $s_e$, an episode executes low-level robot actions $a_i$ until task success or termination, yielding
\refstepcounter{equation}\label{eq:dataset}
$\xi_e=(s_i,a_i)_{i<T_e},\qquad
S_e=\mathbf{1}\!\left[\mathrm{Succ}_{\ell}(s_{T_e})\right],\qquad
\mathcal{D}_{\mathrm{ours}}=\{\xi_e:S_e=1\}$~(\theequation).
Robot control runs at 20\,Hz, while the foundation model acts only at sparse decision points. Its parameters remain fixed throughout collection, and each model call is stateless; previous attempts can affect subsequent episodes only through the explicit experience mechanism introduced in Section~\ref{sec:experience}. Successful executions are recorded in the environment's native action representation and form the demonstration dataset used for downstream policy learning.

Because unsuccessful attempts also consume expensive model inference, we evaluate collection efficiency using cost per demonstration
\refstepcounter{equation}\label{eq:cost_of_pass}
$J=\frac{\mathbb{E}[C]}{P(S=1)}
=\frac{\mathbb{E}[N]\,\bar c}{P(S=1)},
\qquad
\bar c=\frac{\mathbb{E}[C]}{\mathbb{E}[N]}$~(\theequation).
\looseness=-1 where $N$ is the number of API requests in an attempt, $C$ is total list-price cost,
and $\bar c$ is the call-weighted average cost per request. We use $J$ as an evaluation metric rather than an objective optimized directly by the method, and also report requests, tokens, and wall-clock collection time.

\CAPEX combines three mechanisms, illustrated in Fig.~\ref{fig:overview}. First, the VLM acts through a structured interface in which a single model decision can specify a temporally extended robot plan (Section~\ref{sec:waypoint_interface}). 
Second, execution outcomes from previous rollouts self-supervise a confidence calibrator, determining how many proposed waypoints should executed before the robot observes again and replans (Section~\ref{sec:calibrated_commitment}). Third, prior attempts are summarized into bounded cross-episode state that provides the VLM with execution reliability and examples of the robot's previous successful episodes (Section~\ref{sec:experience}).

\subsection{Acting Through a Waypoint Interface}
\label{sec:waypoint_interface}

Querying the VLM after every small robot motion couples collection directly to model latency and inference cost. \CAPEX instead separates semantic visuospatial reasoning from low-level control. At model decision $t$, the VLM receives the task instruction, current observation $o_t$, memory of recent actions $m_t$, any points grounded during the current decision $L_t$, and the cross-episode context $\Gamma_e$. It then proposes a variable-length plan:
\refstepcounter{equation}\label{eq:plan}
$P_t=\big((x_k,R_k,g_k,q_k)\big)_{k=1}^{K_t}
\sim \pi\!\left(\cdot\mid \ell,o_t,h_t,L_t,\Gamma_e\right)$~(\theequation). %
where $x_k$ is a target end-effector position, $R_k$ its orientation, $g_k$ a gripper command, and $q_k\in[0,1]$ the model's stated probability that waypoint $k$ will execute reaches its target without an execution stop, conditional on the preceding waypoints reaching their targets. The model may instead issue a bounded mobile-base motion when repositioning is required. $o_t$ contains multi-camera RGB views, robot state, remaining execution budget, and recent controller outcomes.

\textbf{Tool interface.}
In addition to proposing robot actions, the VLM may invoke a bounded, extensible set of environment-grounding tools that provide reliably computed task-relevant information. Our current toolbox includes \texttt{locate\_pixel}, which maps a selected image pixel to its corresponding 3D position in the current camera view.

\subsection{Calibrated Plan Execution}
\label{sec:calibrated_commitment}

Multi-waypoint plans reduce foundation-model queries only if the robot can commit to actions proposed before observing their consequences. \CAPEX therefore asks the VLM to attach a confidence $q_k$ to every waypoint, interpreted as the probability that waypoint $k$ reaches its target without a stall or contact stop, conditional on all preceding waypoints reaching theirs. Rather than using these self-reported probabilities directly, we calibrate them from the robot's execution outcomes and use the resulting probabilities to determine how much of each plan is executed before observing again.

\textbf{Self-supervised reach labels.}
Each executed waypoint automatically produces a binary label from the low-level executor,
\begin{equation}
y_k=
\begin{cases}
1, & \text{if waypoint $k$ is reported reached}, \\
0, & \text{if it stalls, is contact-stopped, times out, or otherwise fails}.
\end{cases}
\label{eq:reach_label}
\end{equation}
Calibration requires no human annotation or held-out supervision, since execution outcomes themselves provide the training signal. Waypoints that are deferred or never executed receive no label.

\textbf{Confidence calibration.}
Before episode $e$, \CAPEX fits a lightweight reach predictor on all labeled waypoints from previous attempts of the same task. For waypoint $k$, we construct a feature vector $\phi_k$ from the model's stated confidence together with information about the current motion, gripper/phase state, relation to the most recently grounded point, and position within the plan. 
The calibrated reach probability is
\refstepcounter{equation}\label{eq:calibrated_probability}
$\hat p_k=\sigma\!\left(\mathbf{w}_e^\top\phi_k\right)$~(\theequation),
where $\sigma$ is the logistic function. We estimate $\mathbf{w}_e$ by maximum a posteriori (MAP) logistic regression,
\refstepcounter{equation}\label{eq:calibrator}
$\mathbf{w}_e=\arg\min_{\mathbf{w}}
\sum_{(\phi,y)\in\mathcal{Y}_e}
\left[\log\!\left(1+e^{\mathbf{w}^\top\phi}\right)
-y\,\mathbf{w}^\top\phi\right]
+\frac{1}{2}\left\lVert\mathbf{w}-\mathbf{w}_0\right\rVert_2^2$~(\theequation),
with prior mean
\refstepcounter{equation}\label{eq:calibrator_prior}
$\mathbf{w}_0=(0,1,0,\ldots,0)^\top$~(\theequation).
The stated log-odds $\operatorname{logit}(q_k)$ are one component of $\phi_k$, so this prior is centered on identity calibration: before any execution data are available, Eq.~\ref{eq:calibrated_probability} reduces to $\hat p_k=q_k$. The calibrator is refit once at the start of each episode using outcomes accumulated from earlier attempts.

\textbf{Adaptive execution horizon.}
Given a proposed plan of length $K_t$, \CAPEX executes the longest prefix before the first later waypoint whose calibrated reach probability falls below a fixed threshold $\tau$:
\refstepcounter{equation}\label{eq:commit_rule}
$n_t=\min\!\left(
\left\{k\in\{2,\ldots,K_t\}:\hat p_k<\tau\right\}
\cup\{K_t+1\}
\right)-1,\qquad \tau=0.5$~(\theequation).
Every model decision executes at least one waypoint and provides an execution label. Once a later waypoint falls below threshold, that waypoint and the remaining tail are deferred rather than queued for execution. The robot obtains a new observation and asks the VLM to replan from the resulting state. The confidence threshold is a tunable parameter we decide from experience; it reflects a trade-off between replanning frequently with the VLM and acting on waypoints that are predicted to be unreliable. Details are included in Section~\ref{sec:experiments}.

\subsection{Learning from Experience}
\label{sec:experience}

\CAPEX learns from experience accumulated across previous collection attempts. During execution, the robot observes which motions succeed, where execution fails, and which strategies complete the task. \CAPEX records these outcomes and summarizes them into bounded cross-episode state that can be used as experience to inform future episodes.
\refstepcounter{equation}\label{eq:experience_context}
$\Gamma_e=\big(\mathbf{w}_e,\mathcal{M}_e\big),\qquad
\mathcal{M}_e=\big(\mathcal{B}_e,\mathcal{R}_e\big)$~(\theequation).
where $\mathbf{w}_e$ parameterizes the confidence calibrator used to determine the execution horizon. The model-facing context $\mathcal{M}_e$ contains a phase-level reliability summary $\mathcal{B}_e$ and up to two of the robot's previous successful episodes $\mathcal{R}_e$. Thereby, prior experience affects future behavior through two distinct channels: $\mathbf{w}_e$ determines \emph{how far} the robot executes a proposed plan before observing again, while $\mathcal{M}_e$ can influence \emph{what} plan the VLM proposes.

\textbf{Execution calibration.}
All labeled waypoints from previous attempts are used to train the regression model described in Section~\ref{sec:calibrated_commitment}, which calibrates VLM generated reach probabilities. The resulting calibrated reach probabilities $\hat p_k$ determine the execution horizon.

\textbf{Phase-level reliability.}
\CAPEX summarizes which stages of the task have historically executed reliably. Successful trajectories are segmented at gripper events and matched across episodes into recurring manipulation phases. For each phase, the system aggregates execution statistics such as the typical number of proposed waypoints and the fraction of executed waypoints that reached their targets; failed attempts additionally identify phases in which stalls or execution stops occurred. These statistics are converted into a compact textual summary $\mathcal{B}_e$ and included in the VLM context for the next episode. The summary contains no waypoint coordinates or actions. Instead, it indicates which stages have historically been reliable or failure-prone, allowing the VLM to account for prior execution experience when proposing a new plan.

\textbf{Own-success references.}
\CAPEX additionally provides the VLM with up to two previous successful episodes of the same task, selected from different initial states. Each reference includes the episode's initial observations, grounded scene points, executed waypoint sequence, and executor outcomes. When a waypoint can be associated with a grounded scene point, its position is represented relative to that point rather than only in absolute robot coordinates. For a new episode, the VLM is instructed to re-ground the corresponding points in the current scene and use the previous successful episode as a reference when constructing a new plan. These references can therefore provide both an example of a successful manipulation strategy and geometric information about how actions were positioned relative to task-relevant scene elements.

\textbf{Bounded non-parametric adaptation.}
The experience may grow overtime, but the state of each new episode remains bounded. The controller retains a fixed-dimensional calibrator, while the model-facing context contains at most eight reliability-phase entries and two successful references. Once the reference budget is filled, additional collection does not cause the VLM context to grow without bound; all cross-episode adaptation occurs through the explicit state in Eq.~\ref{eq:experience_context}.

\CAPEX thus adapts through experience with the VLM fixed. Previous execution outcomes affect both \emph{how future plans are generated} through phase summaries and successful references, and \emph{how much of those plans is trusted} through the calibrated commitment rule. Successful trajectories simultaneously become part of this in-context experience and the demonstration dataset exported for downstream policy learning.

%% file: sections/4_exp.tex
\section{Experiments}
\label{sec:experiments}

\begin{figure}[t]
\centering
\includegraphics[width=\linewidth]{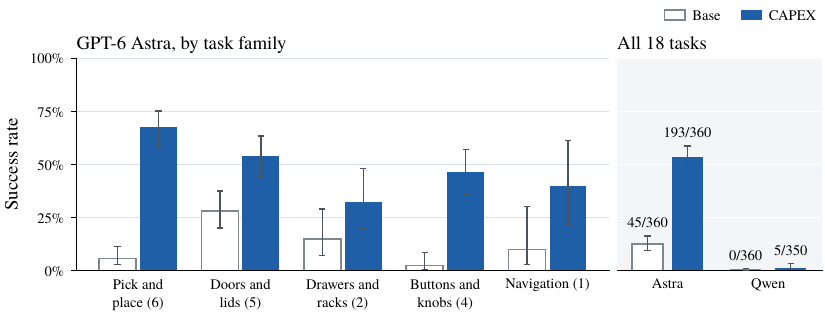}
\caption{Autonomous collection success of \textsc{VLM(Base)}  and \textsc{VLM}+\textsc{CAPEX} on the same 20 recreated human-demonstration starts per task.
Left: GPT-6 Astra, pooled by RoboCasa task family (120, 100, 40, 80 and 20 starts).
Right: all 18 tasks for both backbones.
Whiskers are 95\% Wilson intervals.
\textsc{Astra}+\textsc{CAPEX} outperforms \textsc{Astra(Base)} in 14 of the 18 tasks, 11 of them significantly ($p<0.05$, McNemar).}
\label{fig:collection_success}
\end{figure}

In our experiments we measure the ability of \CAPEX to
(1) efficiently collect successful robot demonstrations, (2) improve subsequent collection using experience from previous executions , and (3) generate demonstrations that provides useful supervision for downstream robot policies across simulation and real-world. We find that \CAPEX collects \(4.3\times\) as many successful demonstrations (Figure~\ref{fig:collection_success}) while reducing cost per success by 80\% in RoboCasa, with prior successful experience further improving collection efficiency. These gains extend to real world deployment: on a bimanual YAM robot, \CAPEX increases autonomous collection success from 45\% to 95\%, while on a Franka Panda, policies trained on \CAPEX demonstrations match or exceed their human-trained counterparts in five of six task--policy comparisons.

\subsection{Experimental Setup}
\label{sec:experimental_setup}

We evaluate autonomous demonstration collection on 18 atomic RoboCasa tasks from the Robocasa365 target dataset \citep{nasiriany2026robocasa365}. The tasks span navigation, articulated-object interaction, and pick-and-place manipulation. For each task, we recreate the initial states of the first 20 official human demonstrations and use their recorded task instructions, yielding 360 paired evaluation episodes. All methods are evaluated on the same starts in the same order.

We mainly collect demonstrations with two VLMs: GPT-6 Astra and Qwen3.8-27B~\citep{qwen38}. For each model, we compare two collection conditions. \textsc{VLM(Base)}  receives an interface-only prompt and predicts a single robot waypoint or base motion per model request. \textsc{VLM}+\textsc{CAPEX} uses multi-waypoint planning, execution calibration, and cross-episode experience as described in Section~\ref{sec:method}. 

For evaluation, we primarily report task success together with the resources required to obtain successful demonstrations: the number of model requests, wall-clock collection time, and API cost at list prices. Cost per successful demonstration is computed as the total API cost of all attempts divided by the number of successful demonstrations. Comparisons of success on matched RoboCasa starts use exact McNemar tests, while paired execution statistics use two-sided sign tests. Unless otherwise specified, costs use API list prices as of September 22, 2026.

\subsection{Autonomous Demonstration Collection}
\label{sec:collection_results}

We found that \textbf{\textsc{Astra}+\textsc{CAPEX} substantially increases successful collection while reducing the cost of each successful demonstration}, compared with \textsc{Astra(Base)}.
Across the 360 matched starts, \textsc{Astra}+\textsc{CAPEX} succeeds on 193 episodes (53.6\%), compared with 45 (12.5\%) for \textsc{Astra(Base)}, a \(4.3\times\) increase in the number of successful demonstrations. This higher success rate translates directly into collection efficiency. The list-price API cost per successful demonstration decreases from \$9.15 to \$1.83, while model requests per success decrease from 244 to 37 and wall-clock collection time from 39.7 to 6.6 minutes (Table~\ref{tab:collection_efficiency}).

The difference is not explained simply by making individual attempts cheaper. \textsc{Astra}+\textsc{CAPEX} costs \$0.98 per attempted episode compared with \$1.14 for \textsc{Astra(Base)}  and uses 19.6 versus 30.3 model requests per attempt. Rather, the largest efficiency gain comes from converting substantially more attempted episodes into usable demonstrations.

\begin{table}[!b]
\centering
\caption{Autonomous collection over the same starts for both backbones.
Success is over all scorable starts with 95\% Wilson intervals.
Cost, model requests and wall-clock time include every attempt and are divided by the number of successful demonstrations;
$\infty$ marks zero successes and --- marks no API price (Qwen runs locally).
$\Delta$: \CAPEX{} minus Base (percentage points for success). Bold: better within each backbone.}
\label{tab:collection_efficiency}
\small
\begin{tabular}{lcccc}
\toprule
Method & Success $\uparrow$ (95\% CI) & \$ / Success $\downarrow$ & Requests / Success $\downarrow$ & Min.\ / Success $\downarrow$ \\
\midrule
\textsc{Astra (Base)} & 45/360~~12.5\% (9.5--16.3) & 9.15 & 244 & 39.7 \\
\textsc{Astra}+\textsc{CAPEX} & \textbf{193/360~~53.6\% (48.4--58.7)} & \textbf{1.83} & \textbf{37} & \textbf{6.6} \\
\midrule
\textsc{Qwen (Base)} & 0/360~~0.0\% (0.0--1.1) & --- & $\infty$ & $\infty$ \\
\textsc{Qwen}+\textsc{CAPEX} & \textbf{5/350~~1.4\% (0.6--3.3)} & --- & \textbf{4{,}819} & \textbf{2{,}332} \\
\bottomrule
\end{tabular}
\end{table}

Alternatively, \textsc{Qwen}+\textsc{CAPEX} yields 5/350 successful demos and \textsc{Qwen(Base)} yielded 0/360 (Table~\ref{tab:collection_efficiency}). This suggests that the collection performance currently largely depends on the foundation model. However, since the capabilities of general-purpose foundation models are promptly improving, we anticipate that most models will work with our framework in the near future.

\subsection{Learning from Experience}
\label{sec:experience_results}

\begin{figure}[t]
\centering
\includegraphics[width=\linewidth]{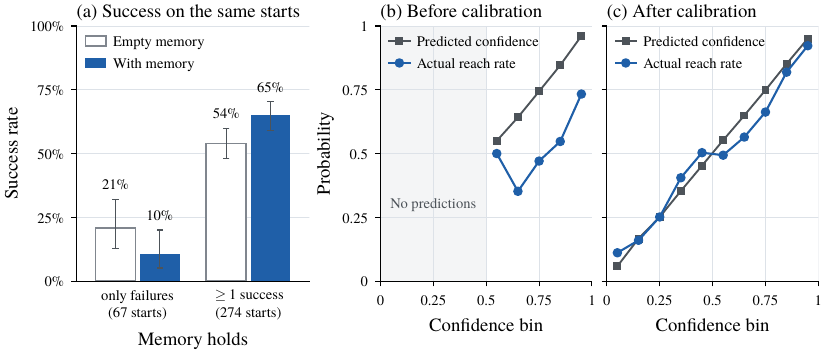}
\caption{Learning from experience (GPT-6 Astra).
(a)~Success on the same starts with memory and with memory cleared, split by whether the memory already holds a successful execution; whiskers are 95\% Wilson intervals.
(b,\,c)~Waypoint confidence before and after calibration, over the same 8{,}458 executed waypoints in ten equal-width confidence bins: for each bin, the mean predicted confidence and the actual share of waypoints that reached their target.
Before calibration, the VLM never predicts below 0.5 and its predictions sit well above the actual reach rate; after calibration the two lines nearly coincide (ECE 0.24 $\rightarrow$ 0.05, AUROC 0.57 $\rightarrow$ 0.75).}
\label{fig:experience}
\end{figure}

We next evaluate whether cross-episode experience improves autonomous collection, and which aspects of prior experience account for the gains. Here, we have three major findings:

\textbf{1) Successful prior experience improves collection on the same starts.}
We isolate memory by rerunning \CAPEX with memory cleared before each episode, holding all other settings fixed and excluding each task's identical first episode. Across 341 paired starts, \textsc{Astra}+\textsc{CAPEX} succeeds on 185 episodes with memory versus 162 without (\(p=0.027\)). The gain is concentrated after at least one prior success: success rises from 148/274 (54.0\%) to 178/274 (65.0\%; \(p=0.0016\)) (Figure~\ref{fig:experience}a). With only failed prior executions, performance instead drops from 14/67 to 7/67 (\(p=0.09\)). Thus, memory helps primarily once the robot has acquired a successful task example.

\textbf{2) Experience makes successful executions shorter and faster.}
On the 124 starts solved under both conditions, memory reduces the median execution from 329 to 273 control steps, from 11 to 8 model requests, and from 124 to 98 seconds of wall-clock time (\(p=0.001\), \(0.004\), and \(<10^{-5}\), respectively). These correspond to reductions of 17\%, 27\%, and 21\%.

\textbf{3) Outcome calibration turns experience into an actionable execution signal.}
\CAPEX uses waypoint confidence to determine how far to execute before replanning. Raw foundation-model confidence is poorly calibrated: 94.9\% of executed waypoints receive confidence $\geq 0.8$ (mean 0.94), while 70\% reach their targets and none fall below the 0.5 commit threshold (Figure~\ref{fig:experience}b). Across 8,458 waypoints, raw confidence yields ECE 0.24 and AUROC 0.57 for predicting waypoint success.

Using outcomes from prior executions, \CAPEX recalibrates confidence before committing. After calibration, predicted confidence tracks the actual reach rate across the full range (Figure~\ref{fig:experience}c): ECE drops from 0.24 to 0.05 and AUROC rises from 0.57 to 0.75. As a result, plan truncation increases from 0.4\% with empty memory to 43.9\% with experience-conditioned calibration. 

\subsection{Downstream Policy Learning in Simulation}
\label{sec:downstream_results}

We next test whether successful \CAPEX rollouts can train deployable policies. We use six tasks with at least 15/20 successful starts and build paired \textsc{Astra}+\textsc{CAPEX} and human datasets from identical initial states, with up to 50 demonstrations per task. We train RoboCasa365-initialized and scratch Diffusion Policies (DP)~\citep{chi2023diffusionpolicy}, plus ACT~\citep{zhao2023learning}, for 20k steps with three seeds, evaluating each for 50 rollouts per seed in the same kitchen configuration.

Across policy classes, \textbf{policies trained on \textsc{Astra}+\textsc{CAPEX} demonstrations approach human-trained performance} after 20k steps (Table~\ref{tab:student-policies}). Across six tasks, success is 0.55 vs.\ 0.61 for fine-tuned DP, 0.29 vs.\ 0.35 from scratch, and 0.28 vs.\ 0.36 for ACT, with \textsc{Astra}+\textsc{CAPEX} outperforming human data on several task--policy pairs.

\begin{table*}[!t]
\centering
\vspace{-1em}
\caption{
Downstream policy success when trained on \CAPEX demonstrations (Ours)
or human teleoperation from the same initial states, with equal dataset sizes
for each task. Values are mean $\pm$ standard deviation over three training
seeds, with 50 evaluation rollouts per seed after 20k training steps.
Bold indicates the higher mean within each task--policy pair.
}
\label{tab:student-policies}
\small
\resizebox{\textwidth}{!}{
\begin{tabular}{lcccccc}
\toprule
& \multicolumn{2}{c}{DP Fine-Tuned}
& \multicolumn{2}{c}{DP From Scratch}
& \multicolumn{2}{c}{ACT} \\
\cmidrule(lr){2-3}
\cmidrule(lr){4-5}
\cmidrule(lr){6-7}
Task (Demos) & Ours & Human & Ours & Human & Ours & Human \\
\midrule
Close Fridge (50)
& \textbf{0.98$\pm$0.02} & 0.89$\pm$0.05
& 0.79$\pm$0.04 & \textbf{0.97$\pm$0.01}
& 0.57$\pm$0.11 & \textbf{0.93$\pm$0.01} \\

Turn On Microwave (46)
& 0.75$\pm$0.05 & \textbf{0.85$\pm$0.08}
& 0.27$\pm$0.09 & \textbf{0.50$\pm$0.05}
& \textbf{0.33$\pm$0.17} & 0.33$\pm$0.08 \\

PnP Counter$\rightarrow$Stove (50)
& 0.25$\pm$0.06 & \textbf{0.29$\pm$0.04}
& \textbf{0.07$\pm$0.01} & 0.01$\pm$0.01
& \textbf{0.29$\pm$0.05} & 0.09$\pm$0.03 \\

Close Toaster Oven (50)
& \textbf{0.81$\pm$0.06} & 0.77$\pm$0.06
& 0.51$\pm$0.07 & \textbf{0.54$\pm$0.09}
& 0.38$\pm$0.07 & \textbf{0.61$\pm$0.11} \\

PnP Toaster$\rightarrow$Counter (33)
& 0.19$\pm$0.05 & \textbf{0.64$\pm$0.05}
& 0.05$\pm$0.06 & \textbf{0.06$\pm$0.02}
& 0.09$\pm$0.03 & \textbf{0.19$\pm$0.08} \\

PnP Sink$\rightarrow$Counter (35)
& \textbf{0.31$\pm$0.05} & 0.19$\pm$0.08
& \textbf{0.05$\pm$0.02} & 0.01$\pm$0.01
& 0.00 & 0.00 \\

\textbf{Average}
& 0.55 & \textbf{0.61}
& 0.29 & \textbf{0.35}
& 0.28 & \textbf{0.36} \\

\bottomrule
\end{tabular}
}
\end{table*}

\begin{figure}[t]
\centering
\includegraphics[width=\linewidth]{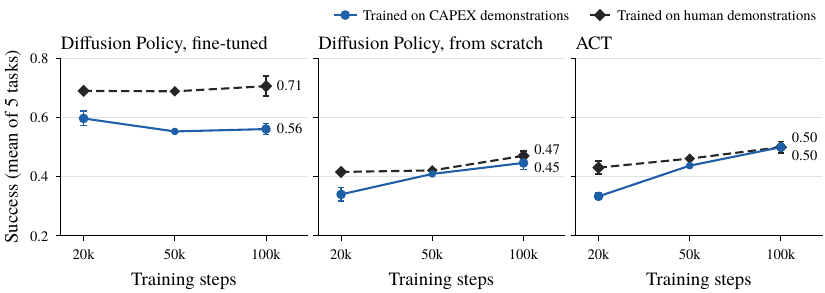}
\caption{Downstream success as a function of training budget, averaged over the five tasks of the extended-budget study.
From left to right: Diffusion Policy initialized from the RoboCasa365 checkpoint, Diffusion Policy trained from scratch, and ACT. The gap between \CAPEX{} and human demonstrations largely closes with additional optimization for policies trained from scratch.}
\label{fig:student_budget}
\end{figure}

We extend training to 100k steps on five tasks to separate demonstration quality from optimization speed. For ACT, mean success rises from 0.33/0.43 with \textsc{Astra}+\textsc{CAPEX}/human data at 20k steps to 0.50/0.50 at 100k; scratch Diffusion Policy similarly improves from 0.34/0.41 to 0.45/0.47 (Figure~\ref{fig:student_budget}). Thus, for policies trained from scratch, much of the early gap narrows with longer training.

This pattern is less apparent on Diffusion Policy initialized from the released RoboCasa365 checkpoint. At 100k steps, success is 0.56 with \textsc{Astra}+\textsc{CAPEX} data versus 0.71 with human demonstrations. Because the checkpoint was pretrained on human data, this comparison favors the human condition. Most of the remaining gap comes from PickPlaceToasterToCounter (0.17 vs.\ 0.70); excluding this task, the averages narrow to 0.66 vs.\ 0.71.

\subsection{Downstream Policy Learning on Physical-Robots}
\label{sec:real_robot}

We evaluate \textsc{Astra(Base)} and \textsc{Astra}+\textsc{CAPEX} on two manipulation deployments:  a bimanual I2RT YAM, and a Franka Panda arm. On YAM, we evaluated placing two objects onto a plate and an in-air bowl handover. Across both tasks, \textsc{Astra}+\textsc{CAPEX} succeeds on 40/42 attempts (95\%), compared with 18/40 (45\%) for \textsc{Astra(Base)}(Table~\ref{tab:physical_robot}). The gap is largest on bowl handover, where success increases from 30\% to 95\%. Per successful demonstration, \CAPEX uses 11.7 vs.\ 37.6 model calls, requires 1.87 vs.\ 4.75 minutes, and reduces API cost from \$1.64 to \$0.65. Thus, the collection gains observed in simulation extend to physical bimanual manipulation.

On a Franka Panda, we train Diffusion Policy and ACT on 20 demonstrations per task collected either by \textsc{Astra}+\textsc{CAPEX} or human teleoperation for \emph{Pick Coke}, \emph{Stack Cups}, and \emph{Stack Blocks}. \CAPEX-trained policies match or exceed their human-trained counterparts in five of six task--policy comparisons (Table~\ref{tab:physical_robot}). Diffusion Policy succeeds on 8/30 vs.\ 6/30 trials and ACT on 4/30 vs.\ 3/30. Together, these results show that \CAPEX can both autonomously collect demonstrations on physical hardware and produce data useful for downstream real-world policy learning.

\begin{table}[t]
\centering
\vspace{-1em}
\caption{
Physical-robot evaluation.
\textbf{Left:} autonomous demonstration collection on the bimanual YAM;
cost is API cost per successful demonstration.
\textbf{Right:} downstream policy success on the Franka Panda;
each entry reports successes over 10 physical trials.
}
\label{tab:physical_robot}
\small
\setlength{\tabcolsep}{4pt}

\begin{tabular}[t]{lcccc}
\multicolumn{5}{c}{\textbf{YAM: Autonomous Collection}} \\[0.35em]
\toprule
& \multicolumn{2}{c}{Success}
& \multicolumn{2}{c}{\$ / Success $\downarrow$} \\
\cmidrule(lr){2-3}
\cmidrule(lr){4-5}
Task & Base & CAPEX & Base & CAPEX \\
\midrule
Bread + Corn
& 12/20 & \textbf{20/21}
& 1.10 & \textbf{0.43} \\
Bowl Handover
& 6/20 & \textbf{20/21}
& 2.73 & \textbf{0.88} \\
\midrule
Overall
& 18/40 & \textbf{40/42}
& 1.64 & \textbf{0.65} \\
\bottomrule
\end{tabular}%
\hfill
\begin{tabular}[t]{lcccc}
\multicolumn{5}{c}{\textbf{Franka: Downstream Policies}} \\[0.35em]
\toprule
& \multicolumn{2}{c}{Diffusion Policy}
& \multicolumn{2}{c}{ACT} \\
\cmidrule(lr){2-3}
\cmidrule(lr){4-5}
Task & Ours & Human & Ours & Human \\
\midrule
Pick Coke
& 4/10 & \textbf{5/10}
& 3/10 & 3/10 \\
Stack Cups
& \textbf{2/10} & 0/10
& \textbf{1/10} & 0/10 \\
Stack Blocks
& \textbf{2/10} & 1/10
& 0/10 & 0/10 \\
\midrule
Overall
& \textbf{8/30} & 6/30
& \textbf{4/30} & 3/30 \\
\bottomrule
\end{tabular}

\end{table}

%% file: sections/5_conclusion.tex
\section{Conclusion}

We presented \CAPEX, an experience-conditioned framework for using multimodal foundation models as autonomous robot demonstrators. Rather than querying the model independently at every control decision, \CAPEX uses execution outcomes to calibrate plan commitment and carries successful experience across collection attempts. Across 18 RoboCasa tasks, this increases successful demonstration collection by \(4.3\times\) while reducing list-price cost per successful demonstration from \$9.15 to \$1.83. The resulting trajectories also support downstream learning: Diffusion Policy and ACT trained on autonomously collected demonstrations approach matched human teleoperation, with the gap largely closing under longer training for policies learned from scratch, and remain effective on physical manipulation tasks. These results suggest a path toward robot learning systems in which foundation models not only act as pretrained controllers, but autonomously generate the experience from which increasingly capable policies can be learned.

\textbf{Limitations and future work.}
\CAPEX changes how a frozen foundation model is used rather than expanding its underlying capabilities. Experience can make a capable demonstrator more effective, but cannot compensate for missing visuomotor competence. Moreover, cross-episode gains emerge primarily after the robot obtains a successful execution; learning better from failures and transferring experience across related tasks remain open directions. Our current objective also prioritizes successful demonstrations rather than their downstream learning value, motivating future collectors that seek trajectories based on how much they improve the student policy. Finally, our physical experiments still require human scene resets and success evaluation. Automating reset, evaluation, and task selection would move toward systems that continually generate and improve their own robot experience.

%% file: appendix/appendix.tex
\renewcommand{\topfraction}{0.9}\renewcommand{\bottomfraction}{0.8}\renewcommand{\textfraction}{0.07}
\renewcommand{\floatpagefraction}{0.8}\setcounter{topnumber}{3}\setcounter{totalnumber}{4}
\lstdefinestyle{rdgprompt}{basicstyle=\renewcommand*{\ttdefault}{lmtt}\ttfamily\scriptsize,columns=fullflexible,
  keepspaces=true,breaklines=true,frame=single,xleftmargin=0pt,xrightmargin=0pt,upquote=true,
  showstringspaces=false,postbreak=\mbox{\textcolor{gray}{$\hookrightarrow$}\space},captionpos=t,
  literate={–}{{\textrm{\textendash}}}1 {—}{{\textrm{\textemdash}}}1 {’}{{'}}1 {→}{{$\rightarrow$}}1 {×}{{$\times$}}1}

\section{Method details}
\label{app:method}

\paragraph{What the model sees and does.}
Every model request is stateless. It contains the system prompt (App.~\ref{app:prompts}), three $512{\times}512$ RGB
images (left and right exterior cameras and the wrist camera), and a text observation: the instruction, the remaining
step budget, the gripper-tip position and orientation in the robot-base frame, the gripper opening with a contact
word (\emph{open}, \emph{closed}, \emph{holding}), and the last six actions with the executor's report on each. The
model never receives object poses or other simulator state. It acts through tools. \texttt{act} gives a plan of up to
24 waypoints, each an absolute or relative tip target in centimetres with an orientation, a gripper command and a
stated confidence $q_k$, or a bounded base motion. \texttt{locate\_pixel}$(c,u,v)$ returns the base-frame position of
pixel $(u,v)$ in camera $c$ from the simulator's depth. A decision allows up to six rounds of tool calls before the
model must act. The base arm uses the same observation and executor, with an \texttt{act} tool of one waypoint and no
confidence, no \texttt{locate\_pixel}, no memory, and it may declare the task done.

\paragraph{Execution and reach labels.}
A conventional executor servos each committed waypoint at 20\,Hz with the operational-space pose controller,
moving at most 10\,cm per waypoint, and then applies its gripper command. A waypoint is reached within 8\,mm and
0.08\,rad; a motion that stops progressing stalls. Control returns to the VLM when the committed prefix finishes or when a
waypoint stalls or makes contact. Reach labels follow the executor's report: $y_k=1$ if the waypoint was reached (a
stall within 1\,cm of the target without contact also counts), $y_k=0$ for any other stall, a contact stop or a
timeout. Waypoints that were deferred or never executed get no label.

\paragraph{Calibrator.}
Table~\ref{tab:app_features} lists the eight features $\phi_k$ of Eq.~\ref{eq:calibrated_probability}. Before each
episode, $\mathbf w_e$ is refit on the labelled waypoints of all earlier attempts at the task (successful or not)
by MAP logistic regression with the prior $\mathcal N(\mathbf w_0, I)$; with no labels, $\hat p_k=q_k$. The first
waypoint of every plan is always executed, so each decision yields at least one label.

\begin{table}[!htbp]
\centering
\footnotesize
\caption{Calibrator features for waypoint $k$ of a plan. Positions in cm in the robot-base frame; $x_0$ is the
current tip position and $x^{\mathrm{loc}}$ the most recent \texttt{locate\_pixel} point of the episode.}
\label{tab:app_features}
\begin{tabular}{@{}cll@{}}
\toprule
$j$ & Feature & $\phi_k[j]$ \\
\midrule
0 & bias & $1$ \\
1 & stated log-odds & $\operatorname{logit}(q_k)$ \\
2 & gripper closed after waypoint $k$ & $\{0,1\}$ \\
3 & gripper phase (gripper commands so far) & $\min(\gamma_k,4)/4$ \\
4 & leg length & $\lVert x_k-x_{k-1}\rVert/10$ \\
5 & vertical leg & $(x_k-x_{k-1})_z/10$ \\
6 & distance to the last located point & $\min(\lVert x_k-x^{\mathrm{loc}}\rVert,40)/20$ \\
7 & position in the plan & $\min(k,4)/4$ \\
\bottomrule
\end{tabular}
\end{table}

\paragraph{Memory.}
After every attempt the robot appends a record: the requested waypoints with their reach labels and executor
outcomes, the located points, the two exterior start images, and the result. Before episode $e$ the records are
summarised into the context of the system prompt.
\begin{itemize}\itemsep1pt
\item The \emph{reliability skeleton} $\mathcal B_e$ segments successful episodes at gripper events into at most
  eight phases and reports, per phase, the typical number of waypoints, how many were reached, and a verdict:
  \emph{chunk-safe} (reach rate $\ge 0.85$ with at least two waypoints typical) or \emph{observe after each waypoint}
  (reach rate $<0.70$). It also lists the phases where failed attempts stalled most. It contains no coordinates
  (Listing~\ref{lst:app_skeleton}).
\item The \emph{own-success references} $\mathcal R_e$ are the task's two most recent successful episodes from
  other starts. Each shows its start images, its located points, and its waypoints written relative to those points,
  with outcomes (Listing~\ref{lst:app_reference}).
\item A \emph{harness settings} line gives the number of labelled waypoints, the calibrator weights $\mathbf w_e$ and
  $\tau$.
\end{itemize}
The context stays bounded as memory grows: the system prompt is 9.8k characters before the first episode and at
most 17.7k afterwards (base arm: 3.9k).

\begin{algorithm}[!htbp]
\caption{\ours{} collection for one task. The base arm skips lines 3--4, 8--10 and 13, and plans one waypoint per request.}
\label{alg:app_collection}
\small
\begin{algorithmic}[1]
\Require instruction $\ell$, starts $s_1,\dots,s_N$, frozen VLM $\pi$, horizon $H$, threshold $\tau=0.5$
\State $\mathcal M\gets\emptyset$, $\mathcal D\gets\emptyset$
\For{$e=1,\dots,N$}
  \State $\mathbf w_e\gets$ MAP fit on all reach labels in $\mathcal M$ \Comment{$\mathbf w_0$ if none}
  \State $\mathcal B_e\gets$ skeleton of $\mathcal M$; \ $\mathcal R_e\gets$ two most recent successes in $\mathcal M$
  \State reset to $s_e$
  \While{not success, fewer than $H$ steps, and within the decision and cost limits}
    \State query $\pi(\cdot\mid\ell,o_t,h_t,L_t,\Gamma_e)$ until it returns a plan $P_t=((x_k,R_k,g_k,q_k))_{k=1}^{K_t}$
      \Comment{\texttt{locate\_pixel} results join $L_t$}
    \State $\hat p_k\gets\sigma(\mathbf w_e^\top\phi_k)$ for $k=1,\dots,K_t$
    \State $n_t\gets\min(\{k\ge2:\hat p_k<\tau\}\cup\{K_t+1\})-1$ \Comment{Eq.~\ref{eq:commit_rule}}
    \State execute waypoints $1,\dots,n_t$; stop early on a stall or contact
    \State label each executed waypoint $y_k$; report every waypoint's outcome in $h_{t+1}$
  \EndWhile
  \State append the episode record to $\mathcal M$
  \State \textbf{if} success \textbf{then} $\mathcal D\gets\mathcal D\cup\{\xi_e\}$
\EndFor
\State \Return $\mathcal D$
\end{algorithmic}
\end{algorithm}

\begin{table}[!htbp]
\centering
\footnotesize
\caption{Collection settings, shared by all tasks. The \qwen{} arms use the same prompts, tools and settings
except the model.}
\label{tab:app_settings}
\begin{tabularx}{\linewidth}{@{}l>{\raggedright\arraybackslash}X@{}}
\toprule
VLM & \astra{} (Responses API), reasoning effort medium, at most 8{,}000 output tokens per request \\
Open-weight VLM & \qwen{} with FP8 weights served by vLLM on one A40; temperature 1.0, top-$p$ 0.95, top-$k$ 20;
  32k-token context \\
Plans & at most 24 waypoints; $\tau=0.5$; the first waypoint always runs \\
Episode limits & horizon $H$ (Table~\ref{tab:app_tasks_results}), 80 decisions, \$25 of API cost \\
Memory & at most 8 skeleton phases and 2 references; all records used for the calibrator \\
Cost & list prices per million tokens: \$10 fresh input, \$1 cached input, \$12.5 cache write, \$50 output
  (reasoning tokens are output). $J$ sums the cost of all attempts, failed ones included, and divides by the
  successes \\
Compute & one job per task and arm: 1 NVIDIA A40 (rendering, and serving \qwen), 8 CPU cores \\
\bottomrule
\end{tabularx}
\end{table}

\FloatBarrier
\section{Prompts}
\label{app:prompts}

Both arms use one system prompt each for all 18 tasks; the task enters only through the instruction in the
observation. Fields in \texttt{<<...>>} are filled per episode from memory
(Listings~\ref{lst:app_skeleton}--\ref{lst:app_reference}). The tool schemas are in the released code.

\begin{lstlisting}[style=rdgprompt,caption={System prompt of \ours{}.},label={lst:app_prompt_ours}]
You control a robot gripper from camera images and measured proprioception. Plan as many waypoints ahead as you can predict with confidence, up to the plan limit. The harness executes your plan in order and returns control to you only when a waypoint stalls or makes contact, or when the plan ends. The environment alone
judges task success, and the episode ends by itself the moment the task is achieved.

Frame and units: x is forward toward the counter, y is left, z is up. The exterior
cameras look roughly along +x, so image-left is +y; the wrist camera looks along the gripper.
Targets are absolute gripper-tip positions in the robot base frame, in centimeters.
Opening is in centimeters, with states "open (empty)", "closed (empty)", or
"holding (opening X cm, finger contact)"; inspect the images before trusting them.
Quaternions are xyzw; Euler roll, pitch, yaw are intrinsic XYZ degrees in the base frame.
"keep" preserves orientation; "down" points along -z and "forward" along +x,
both with fingers closing along base y.

Executor: each displacement is shortened to at most 10 cm and projected into the workspace
box; clamps are reported. A motion stops when the target is reached, when it stalls, or when a
part of the arm other than the hand contacts something. "Stopped by finger contact" means you
are touching the surface at the target: that is normal for pressing, pushing, and grasping.
An "open" is skipped only at the maximum opening with no finger contact; "release" forces open;
"reclose" forces close. locate_pixel(camera, u, v) returns the 3D base-frame position of the
surface at that pixel of any view listed with depth; use it to ground small or distant targets,
and from the wrist view to ground a part you are about to touch.

Plans. An act may carry a chunk: further waypoints executed in order after the first, without a
new observation in between. Each waypoint is clamped to 10 cm from the previous one, so a 30 cm
straight leg is three waypoints. The plan stops at the first waypoint that stalls or is stopped
by contact; a gripper command placed right after a waypoint still runs if that waypoint stalled,
because the fingers may already be on the object. The outcome names which waypoint stopped the
plan and where the tip actually is, and you replan from there. Put everything you can predict from
what you already see into one plan, gripper commands included: straight legs through free space,
hover-descend-close-lift, carry-lower-release-retreat. End the plan only where the outcome of a move
decides what comes next: a first contact with a part you have not seen move, a press or pull that may
or may not move it. A plan that stops early costs nothing but the replanning; a gripper closing
on nothing costs a regrasp.

When base motion is offered, if a target is beyond about 60 cm ahead, 35 cm to the side, or below 10 cm at base height, or the arm keeps striking the fixture around it, use base_forward_cm / base_left_cm / base_turn_deg to put it 35 to 50 cm ahead before reaching. Re-ground the target afterwards because the frame moved. The arm holds its grip while the base moves, so anything the arm cannot pull far enough is pulled by holding it and driving the base backward. A base move cannot carry a chunk. If the instruction is to go to a place or a fixture rather than to manipulate it, the task is the base motion itself: drive toward it in successive base moves, turn to face it, and finish standing in front of it within arm's reach.

How the benchmark scores. Success is only counted when the task state is achieved AND the
gripper has moved well clear of the thing it manipulated: retreat at least 25 cm (up and back)
after manipulating or releasing it, and wait for the episode to end before doing anything else.
Never say done: let the episode end. An articulated part must reach its end stop, so
"open", "closed", "on" and "off" mean fully.

Manipulation. Keep contact while moving a part; move along the arc the part actually travels.
If the part moves but the task does not end, carry it further until no further motion is reported.
If contact is lost, re-establish it before retrying. If the part you touch moves but the task state
does not change, it may have another degree of freedom or you may be on a neighbouring part: try the
other directions first, then change part rather than repeating harder. After two attempts without
movement, change the contact point or approach. Do not repeat a stalled command unchanged.
If a press makes contact but nothing changes, shift 2 to 3 cm and press again before changing approach.
A control under an overhang or beside a larger part is approached from its open side at its own height, not from above.
A free move costs 10 to 20 steps, but contact or a stall can cost 50 to 150.

Judging progress. A thing has moved only if you see it somewhere else than in the previous frame,
or the opening or contact changed: a reached target and "stopped by finger contact" describe
only the gripper. If you cannot name a visible change, say so and test one small deliberate motion.
Before moving a part, identify its moving joint, approximate axis, and direction from the images.
The wrist view helps only when the target is inside it and separable from its background; if the frame
is filled by one surface, a wall, or the thing you stand beside, say that the wrist view does not
show the target and confirm from the exterior views and the measured numbers instead. While carrying
something, a narrowing opening is a slip only if the thing has also stopped moving with the hand: on a
thin object the fingers close a further centimeter while the grasp holds. Lift a few centimeters and
look again; set it down and regrasp lower and squarer only if the opening falls a second time. If no
view contains the thing you are manipulating, say so and use the other evidence: finger contact
persisting through a deliberate 5 cm lift, with the opening steady, is a held object.
Retreat once when you believe the task is done; if the episode has not ended by your next
observation, it is not done: go back and work on it again rather than waiting.

Turning the wrist is far slower than moving it: a quarter turn can cost over two hundred control
steps, a 10 cm move about ten. Choose an approach direction first, ask for that orientation once,
then keep it for several moves; never attach a new orientation to a small corrective step. Ground the
same point from both exterior views and act on it only when they agree within a few centimeters: a
pixel that misses the target lands on whatever is behind it, much further forward at nearly the same
height, so a point further forward than everything else you measured is background.

Reply using act: reasoning (required string, at most 60 words), target_cm (three numbers;
required for arm motion), orientation ("keep", "down", "forward", or three Euler angles; default
"keep"), gripper ("open", "close", "reclose", "release", "keep"; default "keep"), done
(boolean; default false), and optionally chunk: a list of further waypoints, each with target_cm
(absolute) or delta_cm (relative to the previous waypoint), and optional orientation and gripper.
For Codex emit only
{"calls":[{"tool":"act","arguments_json":"{\"reasoning\":\"Hover, descend, grasp, lift.\",\"target_cm\":[40,0,30],\"orientation\":\"down\",\"chunk\":[{\"delta_cm\":[0,0,-9],\"gripper\":\"close\"},{\"delta_cm\":[0,0,12]}]}"}]}.
A locate_pixel call may precede the act in the same reply only if the act does not depend on it.

When a plan closes the gripper on an object, continue the same plan with the lift straight up (about 5 to 10 cm) instead of stopping at the grasp; look again after the lift.

Harness settings:
{"harness": "waypoint", "overlay": "none", "history": 6, "verify": false, "done_gate": false, "done_gate_max": 3, "harness_perception": true, "max_call_displacement_cm": 10.0, "max_model_rounds": 6, "max_motion_steps": 200, "max_gripper_steps": 20, "discrete_step_cm": 2.0, "discrete_coarse_step_cm": 4.0, "discrete_rot_deg": 15.0, "chunk_max": 8, "chunk_transit_only": false, "chunk_budget": null, "plan_max": 24, "plan_confidence": true, "commit_threshold": 0.5, "calibration_labels": <<N_LABELS>>, "calibrator_weights": <<W: 8 floats>>, "base_motion": {"enabled": true, "max_speed_m_s": 0.25, "max_yaw_rate_rad_s": 0.6, "max_translation_m": 0.6, "max_yaw_rad": 1.57}, "perception_tool": "locate_pixel", "workspace_box": <<WORKSPACE_BOX>>, "gripper_max_opening_cm": 8.0, "gripper_open_tolerance_cm": 0.5, "gripper_open_requires_no_finger_contact": true, "gripper_close_threshold_cm": 1.0, "gripper_close_requires_last_close": true, "gripper_close_requires_no_finger_contact": true}

Reference episodes, when shown, are your own earlier successful episodes of this task from other starting configurations (the same kitchen or another): their start images, where you located things, and the waypoints that worked, with what the executor reported. Relate their geometry to the current scene with your own 3D reasoning (where the target is now, what is in the way) and plan for the current scene; do not copy their coordinates.

Give a confidence on every waypoint; low-confidence tails are deferred until you look again.

Reference episodes
<<REFERENCE EPISODES>>

What your past episodes of this task say about open-loop reliability (this is not a path to copy)
<<SKELETON>>

Reply only with tool calls; do not narrate. You may batch perception
(locate_pixel) and memory (remember) calls, followed by at most one robot command.
Calls execute in order. The first robot command ends the reply; all later calls are ignored.
Arguments must be concrete: results of calls in this reply are only available on the next
model round. If a command depends on a new perception result, request perception first.
\end{lstlisting}

\begin{lstlisting}[style=rdgprompt,caption={System prompt of Base \astra{}.},label={lst:app_prompt_base}]
You control a robot gripper from camera images and measured proprioception. Choose
one waypoint per decision, observe its outcome, and replan. The environment alone
judges task success, and the episode ends by itself when the task is achieved.

Frame and units: x is forward toward the counter, y is left, z is up. The exterior
cameras look roughly along +x, so image-left is +y; the wrist camera looks along the gripper.
Targets are absolute gripper-tip positions in the robot base frame, in centimeters;
text positions are rounded to whole centimeters. Opening is in centimeters, with states
"open (empty)", "closed (empty)", or "holding (opening X cm, finger contact)".
These measured states do not certify a grasp or release; inspect the images.
Quaternions are xyzw; Euler roll, pitch, yaw are intrinsic XYZ degrees in the base frame.
"keep" preserves orientation; "down" points along -z and "forward" along +x,
both with fingers closing along base y.

Executor: each displacement is shortened along its straight line to the configured limit
(10 cm by default), then projected into the reset workspace box, whose lower z bound is
the floor. Clamp feedback and the achieved pose are reported. Motions stop when the
target is reached, when they stall, or when a non-finger part of the robot contacts
something. An "open" is skipped only within 0.5 cm of the maximum opening with no finger
contact; a "close" is skipped only below 1 cm when the last gripper command was close.
"reclose" forces close; "release" forces open.

Reply using act: reasoning (required string, at most 60 words), target_cm (three numbers; required for arm motion), orientation ("keep", "down", "forward", or three Euler angles; default
"keep"), gripper ("open", "close", "reclose", "release", "keep"; default "keep"), and done
(boolean; default false). done ends the attempt without moving. When offered, notes (string; default "", at most 600 characters) is your working notes for your next call: plan, what is established, what to try next; returned to you verbatim.
When offered, base_forward_cm, base_left_cm, base_turn_deg default to 0 and specify forward/left displacement and left-positive yaw in the starting base frame; any nonzero value selects only a base move (target_cm omitted or ignored), followed by gripper; caps are 60 cm combined translation and about 90 deg yaw (1.57 rad) unless configured otherwise.
For OpenAI use function calling. For Codex emit only
{"calls":[{"tool":"act","arguments_json":"{\"reasoning\":\"Move above the object.\",\"target_cm\":[40,0,90],\"orientation\":\"down\"}"}]}.
Use memory tools only when offered.

Harness settings:
{"harness": "waypoint", "overlay": "none", "history": 6, "verify": false, "done_gate": false, "done_gate_max": 3, "harness_perception": false, "max_call_displacement_cm": 10.0, "max_model_rounds": 6, "max_motion_steps": 200, "max_gripper_steps": 20, "discrete_step_cm": 2.0, "discrete_coarse_step_cm": 4.0, "discrete_rot_deg": 15.0, "base_motion": {"enabled": true, "max_speed_m_s": 0.25, "max_yaw_rate_rad_s": 0.6, "max_translation_m": 0.6, "max_yaw_rad": 1.57}, "perception_tool": null, "workspace_box": <<WORKSPACE_BOX>>, "gripper_max_opening_cm": 8.0, "gripper_open_tolerance_cm": 0.5, "gripper_open_requires_no_finger_contact": true, "gripper_close_threshold_cm": 1.0, "gripper_close_requires_last_close": true, "gripper_close_requires_no_finger_contact": true}

Reply only with tool calls; do not narrate. You may batch perception
(locate_pixel) and memory (remember) calls, followed by at most one robot command.
Calls execute in order. The first robot command ends the reply; all later calls are ignored.
Arguments must be concrete: results of calls in this reply are only available on the next
model round. If a command depends on a new perception result, request perception first.
\end{lstlisting}

\begin{lstlisting}[style=rdgprompt,caption={Reliability skeleton for PickPlaceCounterToCabinet before its 16th
episode, as it appears in the system prompt.},label={lst:app_skeleton}]
What your past episodes of this task say about open-loop reliability (this is not a path to copy)
Task skeleton for PickPlaceCounterToCabinet from your 15 past episodes (8 successes, 7 failures). Only how reliably each phase executed open-loop is listed; plan the waypoints themselves from the current views.
Phase 1 (open#0): 1 waypoints typical (range 1-1), reached 8/8. chunk with a stop on stall.
Phase 2 (close#1): 4 waypoints typical (range 4-6), reached 32/37; 0 of 2 stalls were followed by a re-try at the same point. chunk-safe.
Phase 3 (release#2): 12 waypoints typical (range 11-13), reached 30/48; 0 of 16 stalls were followed by a re-try at the same point. observe after each waypoint.
Phase 4 (keep#3): 3 waypoints typical (range 2-3), reached 6/7; 0 of 1 stalls were followed by a re-try at the same point. chunk-safe.
Failures spent most stalls in: close#3 (2), close#1 (2), release#4 (1).
Successful episodes used 23 waypoints (range 19-33) and 348 simulator steps.
\end{lstlisting}

\begin{lstlisting}[style=rdgprompt,caption={An own-success reference, as it appears in the system prompt
(shown to CloseFridge episode 5, Fig.~\ref{fig:overview}).},label={lst:app_reference}]
Reference episode 2 (same kitchen layout), success in 3 calls
Located points (base cm): L1 [17,-40,54], L2 [45,-41,-63], L3 [47,-41,60], L4 [132,-38,59]
Waypoints: base cm = located point + relative cm, then what the executor reported.
Call 2: [16,-1,63]=L1+[-1,39,9] forward close reached | [8,-5,63]=L1+[-9,35,9] stalled 1.6 cm from target | [8,-15,63]=L1+[-9,25,9] not executed | [8,-25,63]=L1+[-9,15,9] not executed | [8,-35,63]=L1+[-9,5,9] not executed | [8,-45,63]=L1+[-9,-5,9] not executed | [16,-49,63]=L1+[-1,-9,9] not executed | [20,-40,63]=L1+[3,0,9] not executed (deferred: low confidence)
Call 3: [8,-15,65]=L1+[-9,25,11] reached | [8,-25,65]=L1+[-9,15,11] reached | [8,-35,65]=L1+[-9,5,11] reached | [8,-45,65]=L1+[-9,-5,11] reached | [16,-49,65]=L1+[-1,-9,11] reached | [20,-40,65]=L1+[3,0,11] stalled 4.6 cm from target
Call 4: [19,-32,66]=L1+[2,8,12] stalled 1.1 cm from target | [23,-23,66]=L1+[6,17,12] stalled 1.5 cm from target | [28,-15,66]=L1+[11,25,12] reached | [35,-8,66]=L3+[-12,33,6] reached | [43,-3,66]=L3+[-4,38,6] reached | [52,0,66]=L3+[5,41,6] terminal
\end{lstlisting}

\begin{table}[!htbp]
\centering
\footnotesize
\caption{The first plan of CloseFridge episode 5 (Fig.~\ref{fig:overview}). The calibrator lowers the stated
confidences; the commit rule executes waypoints 1--6 and defers 7--9, the first of which falls below $\tau=0.5$.}
\label{tab:app_example_plan}
\begin{tabular}{@{}lccccccccc@{}}
\toprule
Waypoint $k$ & 1 & 2 & 3 & 4 & 5 & 6 & 7 & 8 & 9 \\
\midrule
Stated $q_k$ & 0.90 & 0.70 & 0.90 & 0.90 & 0.90 & 0.80 & 0.65 & 0.65 & 0.70 \\
Calibrated $\hat p_k$ & 0.88 & 0.54 & 0.85 & 0.84 & 0.84 & 0.67 & 0.40 & 0.42 & 0.53 \\
Outcome & reached & reached & reached & reached & reached & stalled & \multicolumn{3}{c}{deferred} \\
\bottomrule
\end{tabular}
\end{table}

\FloatBarrier
\section{Tasks and per-task results}
\label{app:tasks}

Each task uses the first 20 human demonstrations of RoboCasa's target dataset in kitchen layout~7, style~7. Their
initial states and instructions are recreated exactly; their actions are never shown to the model. All arms see
the same starts in the same order, with one attempt per start. An attempt invalidated by an API or simulator error
is rerun once.

\input{appendix/tab_tasks_results.tex}

\FloatBarrier
\section{Downstream policy learning}
\label{app:students}

\paragraph{Data.}
We use the six tasks where \ours{} succeeded on at least 15 of 20 starts (the next tasks reach 13). For each task,
the \ours{} chain continues on further human starts of the same kitchen until it has 50 successes or the starts run
out. Each \ours{} demonstration is paired with the human demonstration recorded from the same start, giving 50, 46,
50, 50, 33 and 35 demonstrations per source for CloseFridge, TurnOnMicrowave, PickPlaceCounterToStove,
CloseToasterOvenDoor, PickPlaceToasterToCounter and PickPlaceSinkToCounter. Both sources are re-rendered at
$256{\times}256$ by RoboCasa's converter, so they share one image pipeline.

\paragraph{Evaluation.}
Each trained policy runs 50 rollouts in the demonstrations' kitchen with target-split objects, and success counts
if the task condition holds at any step. We report the mean and standard deviation over training seeds 0--2. Per
task, the two sources are compared with Fisher's exact test on pooled rollouts; across tasks, with a
Cochran--Mantel--Haenszel test.

\begin{table}[!htbp]
\centering
\footnotesize
\caption{Policy recipes. DP follows RoboCasa365's Diffusion Policy recipe; ACT uses LeRobot's implementation. Both
sources are trained identically.}
\label{tab:app_student_hparams}
\begin{tabularx}{\linewidth}{@{}l>{\raggedright\arraybackslash}X>{\raggedright\arraybackslash}X@{}}
\toprule
 & Diffusion Policy (fine-tuned / scratch) & ACT \\
\midrule
Inputs & \multicolumn{2}{>{\raggedright\arraybackslash}X}{3 RGB views at $256{\times}256$; end-effector position,
  quaternion and gripper joints} \\
Actions & \multicolumn{2}{>{\raggedright\arraybackslash}X}{12-D at 20\,Hz: end-effector deltas, gripper, base,
  control mode} \\
Language & CLIP ViT-L/14 instruction embedding & none \\
Model & ResNet-18 encoders, transformer denoiser; DDPM, 100 steps; predicts 10 steps, executes 8 & ResNet-18,
  transformer, CVAE (latent 32); chunk 20, temporal ensembling \\
Initialisation & RoboCasa365 multitask checkpoint / random & ImageNet backbone \\
Optimiser & AdamW, lr $10^{-4}$, cosine, 500 warm-up steps, EMA & AdamW, lr $10^{-5}$, constant \\
Training & \multicolumn{2}{>{\raggedright\arraybackslash}X}{batch 16; 20k steps (budget study: 50k and 100k);
  seeds 0--2; final weights evaluated} \\
\bottomrule
\end{tabularx}
\end{table}

\FloatBarrier
\section{Real-robot experiments}
\label{app:real}

\subsection{Franka: policies trained on collected demonstrations}
\label{app:franka}

\paragraph{Robot and collection.}
A Franka Emika Panda arm with its parallel-jaw gripper is controlled like the simulated robot. The model chooses
end-effector waypoints, and an executor servos them at 20\,Hz through Deoxys' operational-space controller. The model
receives RGB images at $640{\times}360$ from a fixed exterior Intel RealSense D435 facing the robot and a
wrist-mounted RealSense D435, but no depth.
The three tasks are \emph{Pick Coke} (``grab the coke can and lift it up''), \emph{Stack Cups} (``stack the orange cup
onto the green cup'') and \emph{Stack Blocks} (``stack the blue block on top of the green block''). 
CAPEX data were collected using the same object configurations as the paired human-teleoperated episodes.
The operator judges success, and failed attempts stay in memory. Collection took 83 attempts for 62 successes, at \$0.53 and 13.3 requests per demonstration: Pick Coke 20/35, Stack Cups 20/20, Stack Blocks 20/26. We keep 20
demonstrations per task. The human-trained policies use the 20 human episodes whose placements those demonstrations
reproduce.

\paragraph{Policies.}
ACT and Diffusion Policy are trained with LeRobot for 100k steps at batch size 8, separately on each source, from the
two RGB views ($640{\times}360$) and the 8-D end-effector state (position, quaternion, gripper width).
Each policy runs 10 closed-loop trials per task, and the operator judges success.

\subsection{Bimanual YAM: collection with and without the framework}
\label{app:yam}

\paragraph{Rig.}
Two 6-DoF I2RT YAM arms on fixed bases 60\,cm apart, each with a parallel gripper. A front
camera above the table looks down at the workspace, and each arm carries a wrist camera (RealSense, $640{\times}480$).
The model receives the three RGB images at every decision. One \texttt{act} gives each arm an absolute fingertip
target in the base frame, an orientation and a gripper command. An inverse-kinematics servo executes it at 20\,Hz, at
most 4\,mm and 0.039\,rad per tick, with each move clamped to 10\,cm; the arms move together, then their grippers act.
The rig has no automatic success check: the model ends an attempt with \texttt{done}, and the operator labels it.

\paragraph{Teachers.}
Both teachers are \astra{} (Responses API, medium reasoning effort, stateless requests) with the same robot, cameras,
executor and safety limits (Table~\ref{tab:app_yam_teachers}).

\begin{table}[!htbp]
\centering
\footnotesize
\caption{The two teachers on the YAM rig.}
\label{tab:app_yam_teachers}
\begin{tabularx}{\linewidth}{@{}l>{\raggedright\arraybackslash}X>{\raggedright\arraybackslash}X@{}}
\toprule
 & Base \astra{} & \ours{} \\
\midrule
Prompt & interface only (the two-arm version of the simulation base prompt) & the simulation prompt plus YAM
  embodiment passages (grasp alignment, rim pinching) \\
Per decision & one waypoint & a plan of up to 24 waypoints, each with a stated confidence; committed up to the first
  later waypoint with stated confidence below 0.5 \\
Perception & RGB only & RGB, and \texttt{locate\_pixel} on calibrated depth \\
Memory & none & reliability skeleton and two of the run's own successes with their start images \\
\bottomrule
\end{tabularx}
\end{table}

\paragraph{Tasks and protocol.}
In \emph{bread and corn to plate}, the left arm picks up a toy bread and the right arm a toy corn, and they place them
on a plate one after the other: the bread on the left half first, then the corn on the right half. An attempt is
accepted only if the left grasp succeeds on the first try, the plate does not move and the bread goes first. In
\emph{bowl handover}, the left arm lifts a bowl by its rim, hands it to the right arm in the air, and the right arm
sets it down upright on the right side; the bowl may not touch the table between lift and set-down. Base trial $N$
starts from the arrangement of \ours{} trial $N$: the operator matches the scene to an overlay of that trial's
recorded front image. \ours{} collected until 20 successes
(21 attempts per task); Base \astra{} made 20 attempts per task. Collection time runs from the start of an attempt to its end,
including model latency and robot motion, and excludes resets and labelling.

\begin{table}[!htbp]
\centering
\footnotesize
\caption{Collection on the YAM rig. Success is the operator's label, with a 95\% Wilson interval. Time, model calls
and API cost are totals over all attempts divided by the successes. Both teachers ended every attempt with
\texttt{done}; the operator rejected 22 of Base \astra{}'s 40 claims and 2 of \ours{}'s 42.}
\label{tab:app_yam_results}
\setlength{\tabcolsep}{4pt}
\begin{tabular}{@{}llccccc@{}}
\toprule
Task & Teacher & Success & 95\% CI & Min./success & Calls/success & \$/success \\
\midrule
Bread and corn & Base \astra{} & 12/20 & 39--78\% & 3.33 & 25.7 & 1.10 \\
 & \ours{} & 20/21 & 77--99\% & 1.49 & 7.5 & 0.43 \\
\addlinespace[2pt]
Bowl handover & Base \astra{} & 6/20 & 15--52\% & 7.58 & 61.5 & 2.73 \\
 & \ours{} & 20/21 & 77--99\% & 2.26 & 15.8 & 0.88 \\
\midrule
Both & Base \astra{} & 18/40 & & 4.75 & 37.6 & 1.64 \\
 & \ours{} & 40/42 & & 1.87 & 11.7 & 0.65 \\
\bottomrule
\end{tabular}
\end{table}

\FloatBarrier
\section{Qualitative examples}
\label{app:qualitative}

Figures~\ref{fig:app_franka_rollouts}--\ref{fig:app_filmstrips} show recorded attempts on the Franka, the YAM rig and in RoboCasa.

\input{appendix/fig_franka_rollouts.tex}
\input{appendix/fig_yam.tex}
\input{appendix/fig_filmstrips_compact.tex}

\FloatBarrier

%% file: appendix/tab_tasks_results.tex
\begin{table}[t]
\centering
\scriptsize
\caption{The 18 RoboCasa tasks and per-task collection results on the 20 recreated human-demonstration starts per
task (successes out of 20 unless a denominator is shown, which counts starts with a valid episode). $H$: horizon in
control steps at 20\,Hz. Success criteria summarise RoboCasa's \texttt{\_check\_success}, evaluated after every
control step; joint positions are normalised to $[0,1]$ over the joint range. \$/demo: list-price API cost of all
attempts divided by successes.}
\label{tab:app_tasks_results}
\setlength{\tabcolsep}{2.5pt}
\begin{tabularx}{\linewidth}{@{}l r >{\raggedright\arraybackslash}X rr rr rr@{}}
\toprule
 & & & \multicolumn{2}{c}{\astra} & \multicolumn{2}{c}{\qwen} & \multicolumn{2}{c}{\$/demo} \\
\cmidrule(lr){4-5}\cmidrule(lr){6-7}\cmidrule(lr){8-9}
Task & $H$ & Success criterion & Base & \ours & Base & \ours & Base & \ours \\
\midrule
\multicolumn{9}{@{}l}{\textit{Pick and place}}\\
CoffeeSetupMug & 600 & Mug within 4\,cm (horizontal), 10\,cm (vertical) of dispenser site; gripper $>$25\,cm from object & 3 & 6 & 0 & 0 & 8.14 & 5.80 \\
PnPCounter$\to$Cabinet & 750 & Object bounding box inside cabinet interior; gripper $>$25\,cm from object & 2 & 12 & 0 & 0 & 14.57 & 1.43 \\
PnPCounter$\to$Stove & 600 & Object touches pan, within 7\,cm of its centre (horizontal); gripper $>$25\,cm from object & 0 & 20 & 0 & 3 & --- & 0.41 \\
PnPDrawer$\to$Counter & 750 & Object touches a counter; gripper $>$25\,cm from object & 0 & 13 & 0 & 0 & --- & 1.81 \\
PnPSink$\to$Counter & 900 & Object on plate (contact, within 0.7 plate radius); plate touches counter; gripper $>$25\,cm from object & 0 & 15 & 0 & 0 & --- & 1.89 \\
PnPToaster$\to$Counter & 600 & Object on plate (contact, within 0.7 plate radius); gripper $>$25\,cm from object & 2 & 15 & 0 & 0 & 12.37 & 1.03 \\
\addlinespace[2pt]\multicolumn{9}{@{}l}{\textit{Doors and lids}}\\
CloseBlenderLid & 900 & Lid within 4\,cm of closed pose, tilt $<$7$^\circ$; gripper $>$15\,cm from lid & 0 & 0 & 0 & 0 & --- & --- \\
CloseFridge & 900 & Every fridge door joint $\le$0.005 of its range & 19 & 20 & 0 & 0/13 & 0.77 & 0.31 \\
CloseToasterOvenDoor & 450 & Door joint $\le$0.005 of its range & 0 & 18 & 0 & 0 & --- & 0.53 \\
OpenCabinet & 1050 & Every door joint $\ge$0.90 of its range & 9 & 8 & 0 & 0 & 3.42 & 2.83 \\
OpenStandMixerHead & 450 & Head joint $>$0.99 of its range & 0 & 8 & 0 & 0 & --- & 1.56 \\
\addlinespace[2pt]\multicolumn{9}{@{}l}{\textit{Drawers and racks}}\\
OpenDrawer & 750 & Drawer slid out $\ge$95\% of 0.55$\times$ its depth & 1 & 0 & 0 & 0/19 & 24.67 & --- \\
SlideDishwasherRack & 450 & Rack joint $\ge$0.95 of range (out) or $\le$0.05 (in) & 5 & 13 & 0 & 1/19 & 2.81 & 1.40 \\
\addlinespace[2pt]\multicolumn{9}{@{}l}{\textit{Buttons, knobs, levers}}\\
TurnOffStove & 750 & Named burner's knob within 0.35\,rad of off & 1 & 7 & 0 & 1 & 25.42 & 2.94 \\
TurnOnElectricKettle & 450 & Switch lever pressed to $\ge$0.95 of its range & 0 & 10 & 0 & 0 & --- & 2.71 \\
TurnOnMicrowave & 450 & Start button pressed; gripper then $>$15\,cm from it & 0 & 20 & 0 & 0 & --- & 0.32 \\
TurnOnSinkFaucet & 600 & Handle joint angle in $(0.40,\pi)$\,rad (water on) & 1 & 0 & 0 & 0/19 & 23.92 & --- \\
\addlinespace[2pt]\multicolumn{9}{@{}l}{\textit{Navigation}}\\
NavigateKitchen & 450 & Base within 0.20\,m of goal pose at the fixture; cos(yaw error) $\ge$0.98 & 2 & 8 & 0 & 0 & 2.85 & 1.69 \\
\midrule
\textbf{All} &  &  & 45/360 & 193/360 & 0/360 & 5/350 & 9.15 & 1.83 \\
\bottomrule
\end{tabularx}
\end{table}

%% file: appendix/fig_franka_rollouts.tex
\begin{figure}[!htbp]
\centering
\setlength{\tabcolsep}{0.5pt}
\renewcommand{\arraystretch}{0.5}
\begin{tabular}{r*{6}{c}}
\rotatebox{90}{\scriptsize\,Pick Coke} & \includegraphics[width=0.155\linewidth]{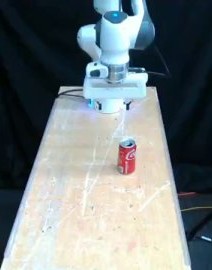} & \includegraphics[width=0.155\linewidth]{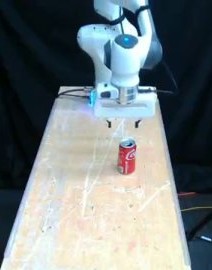} & \includegraphics[width=0.155\linewidth]{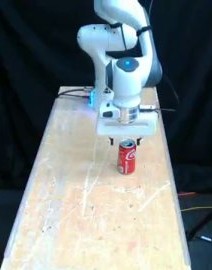} & \includegraphics[width=0.155\linewidth]{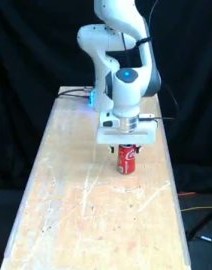} & \includegraphics[width=0.155\linewidth]{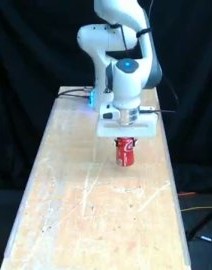} & \includegraphics[width=0.155\linewidth]{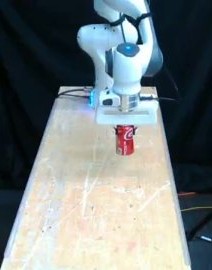} \\[1pt]
\rotatebox{90}{\scriptsize\,Stack Cups} & \includegraphics[width=0.155\linewidth]{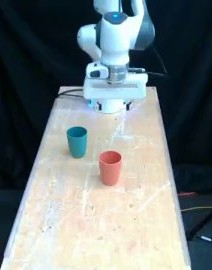} & \includegraphics[width=0.155\linewidth]{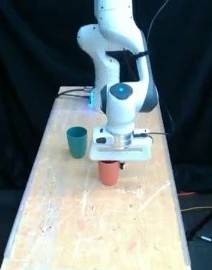} & \includegraphics[width=0.155\linewidth]{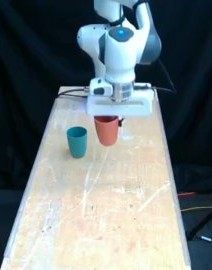} & \includegraphics[width=0.155\linewidth]{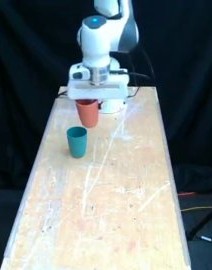} & \includegraphics[width=0.155\linewidth]{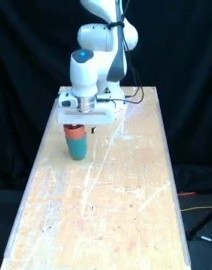} & \includegraphics[width=0.155\linewidth]{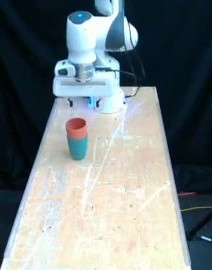} \\[1pt]
\rotatebox{90}{\scriptsize\,Stack Blocks} & \includegraphics[width=0.155\linewidth]{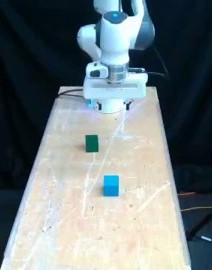} & \includegraphics[width=0.155\linewidth]{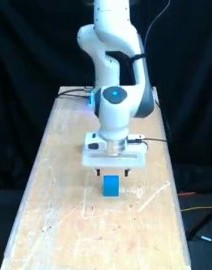} & \includegraphics[width=0.155\linewidth]{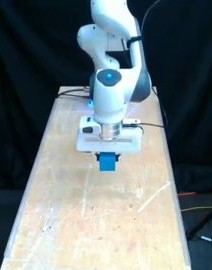} & \includegraphics[width=0.155\linewidth]{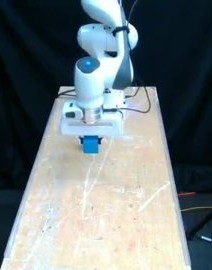} & \includegraphics[width=0.155\linewidth]{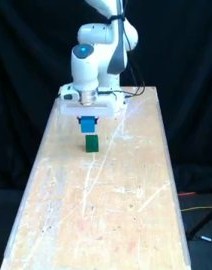} & \includegraphics[width=0.155\linewidth]{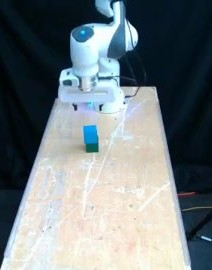} \\[1pt]
\end{tabular}
\caption{Closed-loop rollouts on the Franka of Diffusion Policy trained only on 20 \ours{} demonstrations per task, six evenly spaced frames from the exterior camera (cropped to the workspace). Instructions: \emph{Pick Coke}, ``grab the coke can and lift it up''; \emph{Stack Cups}, ``stack the orange cup onto the green cup''; \emph{Stack Blocks}, ``stack the blue block on top of the green block''. Durations: Pick Coke 9\,s, Stack Cups 34\,s, Stack Blocks 38\,s.}
\label{fig:app_franka_rollouts}
\end{figure}

%% file: appendix/fig_yam.tex
\begin{figure}[!htbp]
\centering
\setlength{\tabcolsep}{0.5pt}
\renewcommand{\arraystretch}{0.5}
\begin{tabular}{r*{6}{c}}
\rotatebox{90}{\scriptsize\,\ours} & \includegraphics[width=0.155\linewidth]{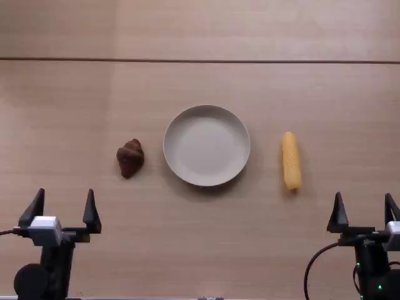} & \includegraphics[width=0.155\linewidth]{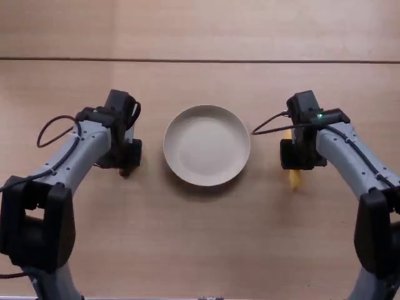} & \includegraphics[width=0.155\linewidth]{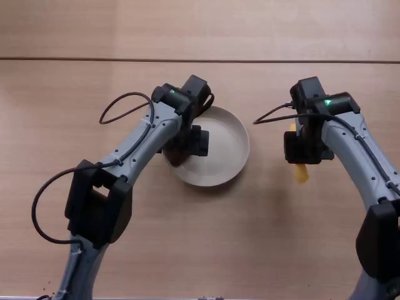} & \includegraphics[width=0.155\linewidth]{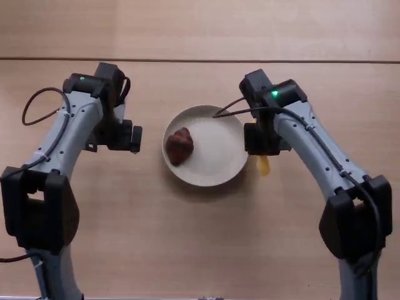} & \includegraphics[width=0.155\linewidth]{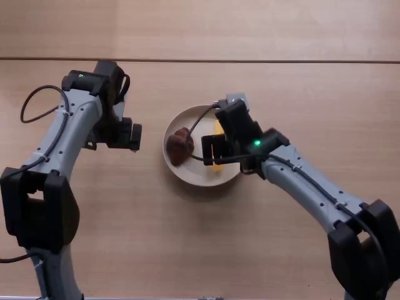} & \includegraphics[width=0.155\linewidth]{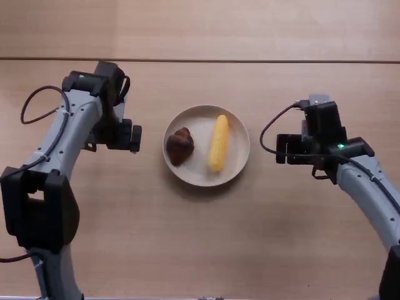} \\
\rotatebox{90}{\scriptsize\,Base} & \includegraphics[width=0.155\linewidth]{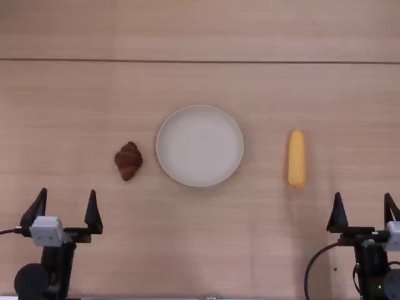} & \includegraphics[width=0.155\linewidth]{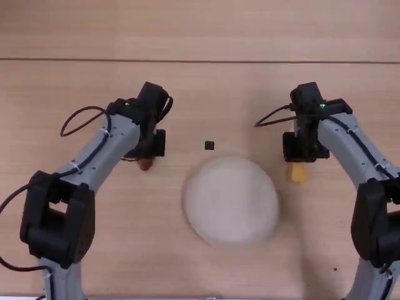} & \includegraphics[width=0.155\linewidth]{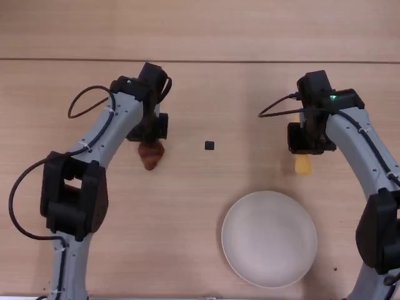} & \includegraphics[width=0.155\linewidth]{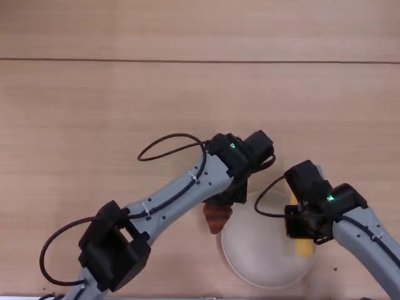} & \includegraphics[width=0.155\linewidth]{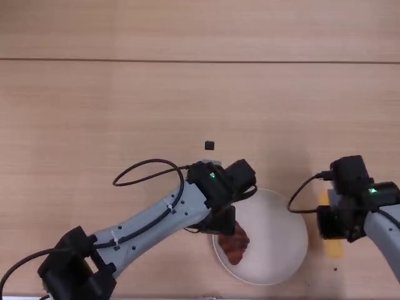} & \includegraphics[width=0.155\linewidth]{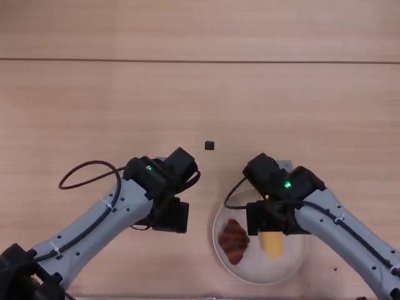} \\[3pt]
\rotatebox{90}{\scriptsize\,\ours} & \includegraphics[width=0.155\linewidth]{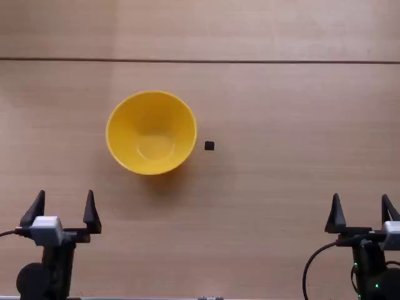} & \includegraphics[width=0.155\linewidth]{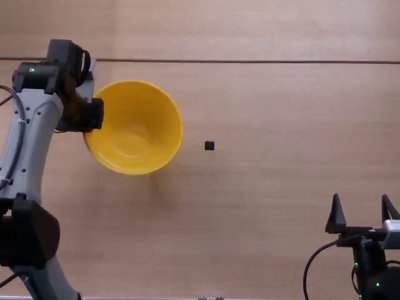} & \includegraphics[width=0.155\linewidth]{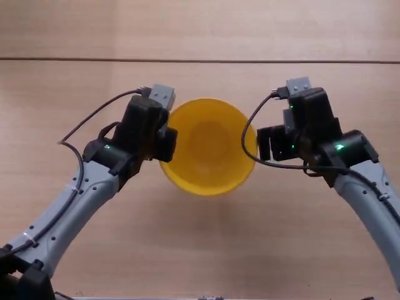} & \includegraphics[width=0.155\linewidth]{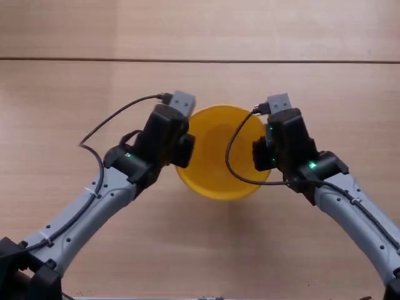} & \includegraphics[width=0.155\linewidth]{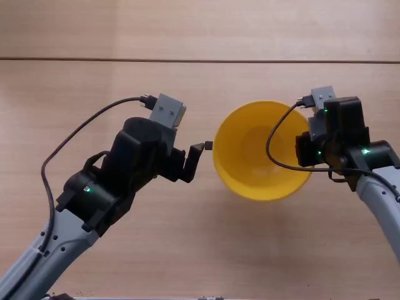} & \includegraphics[width=0.155\linewidth]{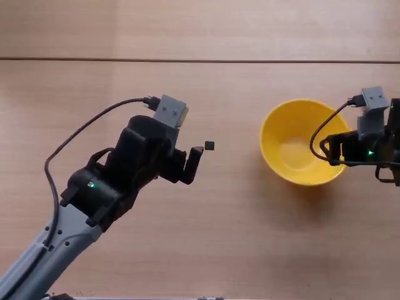} \\
\end{tabular}
\caption{Bimanual YAM, front camera, six evenly spaced frames per attempt. Top two rows: bread and corn to plate from the same start (trial 5). \ours{} places both items without moving the plate (32\,s); Base \astra{} pushes the plate off its position, reports the task done, and the operator labels the attempt a failure (37\,s). Bottom row: bowl handover, \ours{} (trial 0, 49\,s): lift by the rim, hand-over in the air, set-down upright on the right.}
\label{fig:app_yam}
\end{figure}

%% file: appendix/fig_filmstrips_compact.tex
\begin{figure}[!htbp]
\centering
\setlength{\tabcolsep}{0.5pt}
\renewcommand{\arraystretch}{0.5}
\begin{tabular}{r*{6}{c}}
\rotatebox{90}{\scriptsize\,\ours} & \includegraphics[width=0.135\linewidth]{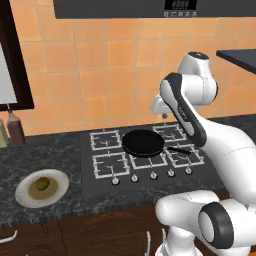} & \includegraphics[width=0.135\linewidth]{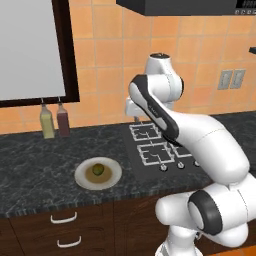} & \includegraphics[width=0.135\linewidth]{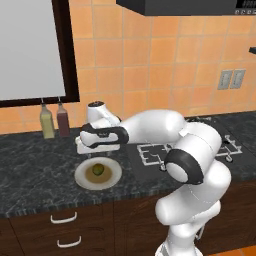} & \includegraphics[width=0.135\linewidth]{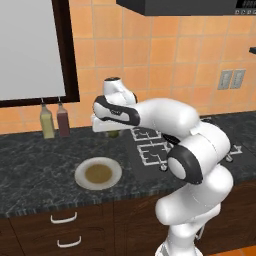} & \includegraphics[width=0.135\linewidth]{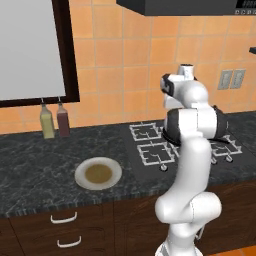} & \includegraphics[width=0.135\linewidth]{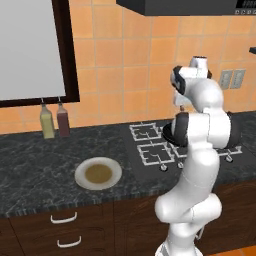} \\
\rotatebox{90}{\scriptsize\,Human} & \includegraphics[width=0.135\linewidth]{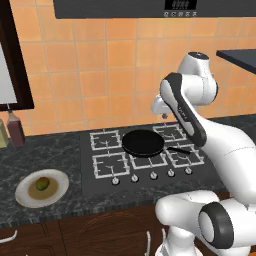} & \includegraphics[width=0.135\linewidth]{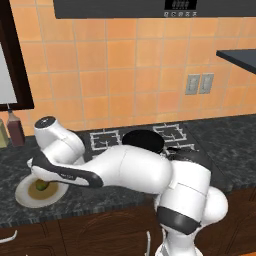} & \includegraphics[width=0.135\linewidth]{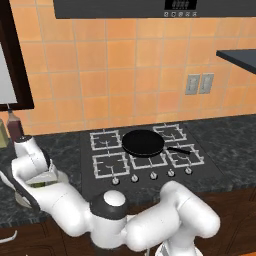} & \includegraphics[width=0.135\linewidth]{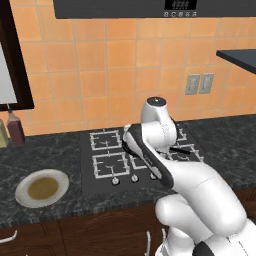} & \includegraphics[width=0.135\linewidth]{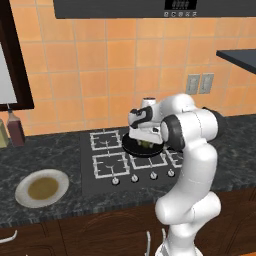} & \includegraphics[width=0.135\linewidth]{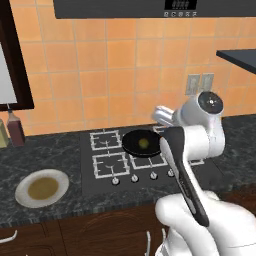} \\
\end{tabular}\\[-1pt]
{\scriptsize \emph{Pick the mango from the plate and place it in the pan.} (PnP Counter To Stove): \ours{} 239 frames (11.9\,s), human 212 frames (10.6\,s).}\\[5pt]
\begin{tabular}{r*{6}{c}}
\rotatebox{90}{\scriptsize\,\ours} & \includegraphics[width=0.135\linewidth]{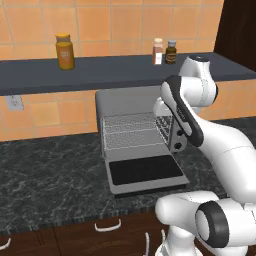} & \includegraphics[width=0.135\linewidth]{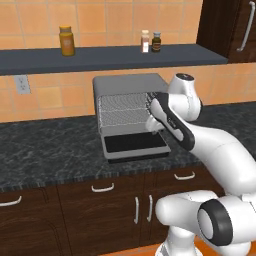} & \includegraphics[width=0.135\linewidth]{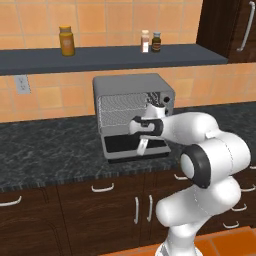} & \includegraphics[width=0.135\linewidth]{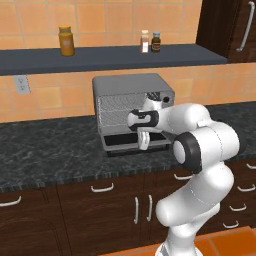} & \includegraphics[width=0.135\linewidth]{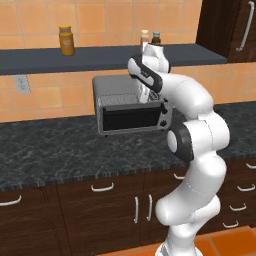} & \includegraphics[width=0.135\linewidth]{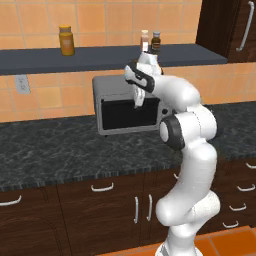} \\
\rotatebox{90}{\scriptsize\,Human} & \includegraphics[width=0.135\linewidth]{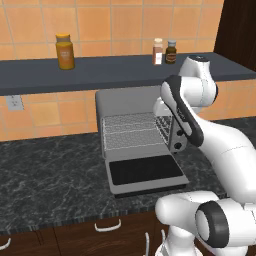} & \includegraphics[width=0.135\linewidth]{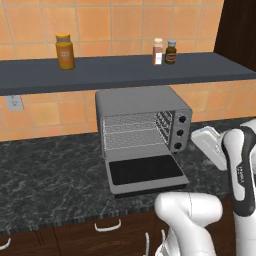} & \includegraphics[width=0.135\linewidth]{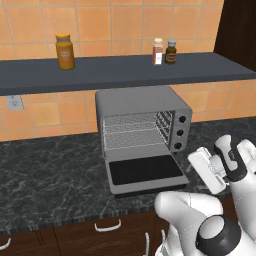} & \includegraphics[width=0.135\linewidth]{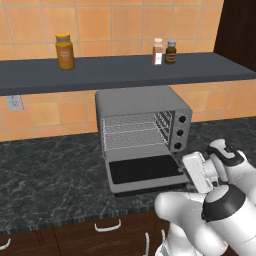} & \includegraphics[width=0.135\linewidth]{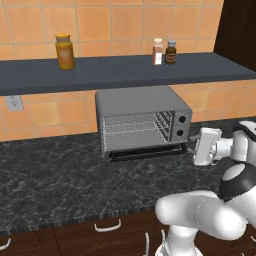} & \includegraphics[width=0.135\linewidth]{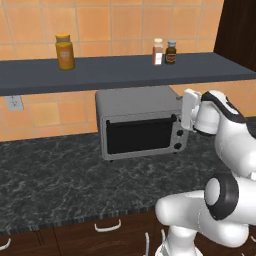} \\
\end{tabular}\\[-1pt]
{\scriptsize \emph{Close the toaster oven door.} (Close Toaster Oven Door): \ours{} 270 frames (13.5\,s), human 178 frames (8.9\,s).}\\[5pt]
\begin{tabular}{r*{6}{c}}
\rotatebox{90}{\scriptsize\,\ours} & \includegraphics[width=0.135\linewidth]{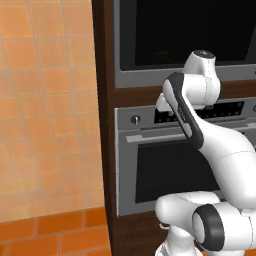} & \includegraphics[width=0.135\linewidth]{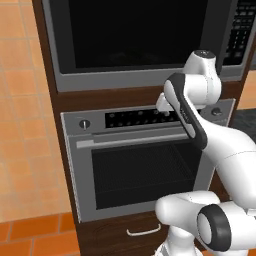} & \includegraphics[width=0.135\linewidth]{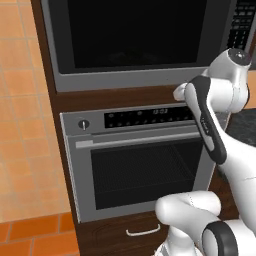} & \includegraphics[width=0.135\linewidth]{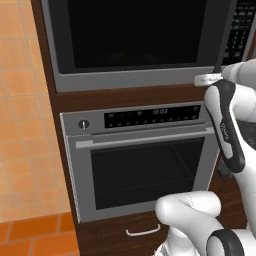} & \includegraphics[width=0.135\linewidth]{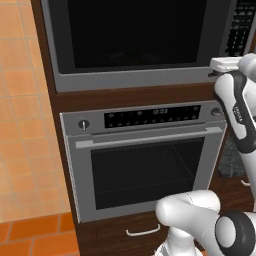} & \includegraphics[width=0.135\linewidth]{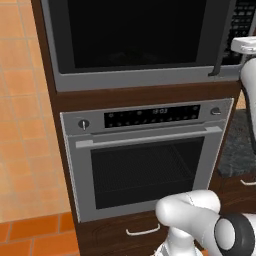} \\
\rotatebox{90}{\scriptsize\,Human} & \includegraphics[width=0.135\linewidth]{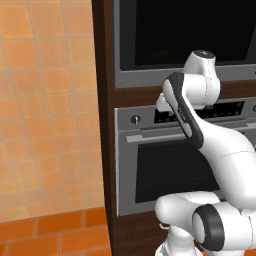} & \includegraphics[width=0.135\linewidth]{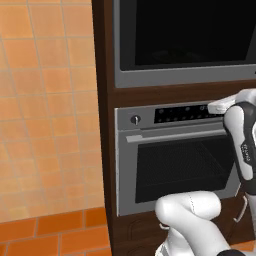} & \includegraphics[width=0.135\linewidth]{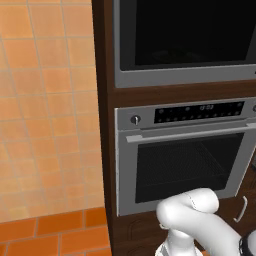} & \includegraphics[width=0.135\linewidth]{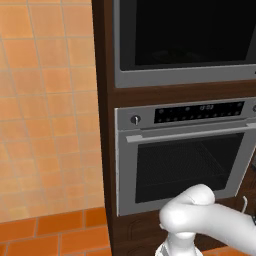} & \includegraphics[width=0.135\linewidth]{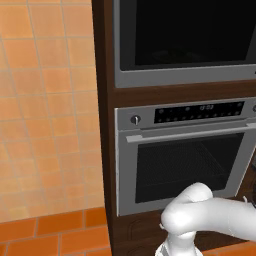} & \includegraphics[width=0.135\linewidth]{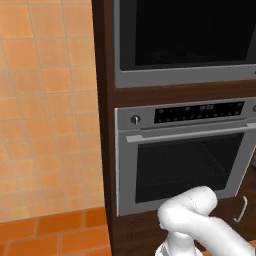} \\
\end{tabular}\\[-1pt]
{\scriptsize \emph{Press the start button on the microwave.} (Turn On Microwave): \ours{} 146 frames (7.3\,s), human 133 frames (6.7\,s).}\\[5pt]
\caption{Demonstrations collected by \ours{} (top rows) and the human teleoperated demonstrations recreated from the same initial states (bottom rows), six evenly spaced frames each; for each task the \ours{} demonstration closest to the task's median length is shown. These pairs are training examples of the downstream policy study (Sec.~\ref{sec:downstream_results}).}
\label{fig:app_filmstrips}
\end{figure}